\documentclass[graybox]{svmult}

\usepackage{type1cm}        
\usepackage{makeidx}         
\usepackage{graphicx}        
\usepackage{multicol}        
\usepackage[bottom]{footmisc}
\usepackage{breakcites}
\usepackage{url}
\usepackage{newtxtext}
\usepackage{newtxmath}       
\usepackage[normalem]{ulem}

\makeindex             
\usepackage{array}
\usepackage{subcaption}
\usepackage{comment}
\usepackage{multirow}
\def\mystrut(#1,#2){ \vrule height #1 depth #2 width 0pt}
\newcolumntype{C}[1]{%
   >{\mystrut(1ex,1.5ex)\centering}%
   m{#1}%
   <{}}
\renewcommand{\arraystretch}{1}

\begin{document}

\newpage

\title*{Analysis of Different Losses for Deep Learning Image Colorization}

\author{Coloma Ballester, Aurélie Bugeau, Hernan Carrillo, Michaël Clément, Rémi Giraud, Lara Raad, Patricia Vitoria}
\authorrunning{C. Ballester, A. Bugeau, H. Carrillo, M. Clément, R. Giraud, L. Raad, P. Vitoria}

\institute{Coloma Ballester (\email{coloma.ballester@upf.edu}) \at Universitat Pompeu Fabra, Barcelona, Spain.
\and Aurélie Bugeau (\email{aurelie.bugeau@labri.fr}) \at Univ. Bordeaux, LaBRI, CNRS UMR 5800, France.
\and Hernan Carrillo (\email{hernan.carrillo-lindado@u-bordeaux.fr}) \at Univ. Bordeaux, LaBRI, CNRS UMR 5800, France.
\and Michaël Clément (\email{michael.clement@labri.fr}) \at Univ. Bordeaux, Bordeaux INP, LaBRI, CNRS UMR 5800, France.
\and Rémi Giraud (\email{remi.giraud@ims-bordeaux.fr}) \at Univ. Bordeaux, Bordeaux INP, IMS, CNRS UMR 5218, France.
\and  Lara Raad (\email{lara.raadcisa@esiee.fr}) \at Université Paris-Est, LIGM (UMR 8049), CNRS, ENPC, ESIEE Paris, UPEM, France.
\and Patricia Vitoria (\email{patricia.vitoria@upf.edu}) \at Universitat Pompeu Fabra, Barcelona, Spain.}

\maketitle

\vspace{-2cm}

\abstract{Image colorization aims to add color information to a grayscale image in a realistic way. Recent methods mostly rely on deep learning strategies. While learning to automatically colorize an image, one can define well-suited objective functions related to the desired color output. Some of them are based on a specific type of error between the predicted image and ground truth one, while other losses rely on the comparison of perceptual properties.
But, is the choice of the objective function that crucial, \emph{i.e.},~does it play an important role in the results? In this chapter, we aim to answer this question by analyzing the impact of the loss function on the estimated colorization results. To that goal, we  review the different losses and evaluation metrics that are used in the literature. We then train a baseline network with several of the reviewed objective functions: classic L1 and L2 losses, as well as more complex combinations such as Wasserstein GAN and VGG-based LPIPS loss.
Quantitative results show that the models trained with VGG-based LPIPS provide overall slightly better results for most evaluation metrics. Qualitative results exhibit more vivid colors when with Wasserstein GAN plus the L2 loss or again with the VGG-based LPIPS.
Finally, the convenience of quantitative user studies is also discussed to overcome the difficulty of properly assessing on colorized images, notably for the case of old archive photographs where no ground truth is available.}

\newpage

\section{Introduction}
\label{sec:intro}

Color is acknowledged to be captured by the human visual system at the first milliseconds.
Color perception allows to highly increase the perceived diversity of real scenes since more than $2$ million colors are identified by humans.
Besides, although humans are interested in color and have used it since the dawn of Humanity, full comprehension of the chromatic aspect of color is still an open problem.
olor images capturing a real scene indeed include both structure information (edges, textures) which is mostly contained in the so-called black-and-white component of the image, and chromatic information which, when added to the achromatic black-and-white component, provides the rich color vision of the scene image.
This achromatic and chromatic dichotomy is also palpable in works of art: artists often slide between drawing strength from the massive richness of the variations on black and white, and exploiting the infinite power of color, even using it as an actor on its own.

Image colorization aims to hallucinate the missing color information of a given grayscale image by, as in the case of learning-based methods, directly learning a mapping from the grayscale to the color information by minimizing a chosen objective function.
The objective function favors the desired properties the estimated colorization should satisfy.
Due to the ill-posed nature of the problem, in most cases, one does not aim to recover the actual ground truth color -- that is, the real color of the actual scene captured in the grayscale image --, but rather to produce a plausible colorization for a human observer.
Accordingly, choosing the right way to train such networks is not trivial.
The network could end up penalizing a good solution far away from the ground truth data or estimating an average of all possible correct solutions. Alternatively, instead of directly learning the per-pixel chrominance information, some methods learn a per-pixel color distribution to, afterward, sample from it the color at each pixel.
In principle, this could encourage the mapping to be one-to-many, which can be desirable.
However, how to properly capitalize and train such networks to account for the different possible solutions having, both, geometric and semantic meaning remains an open problem.

This chapter aims to analyze the influence of the optimized objective function on the results of automatic deep learning methods for image colorization.
Some of the chosen objective functions favor colorization results perceptually as plausible as the associated color ground truth image, no matter the pixel-wise color differences between them, while others aim to recover the ground truth values.
To the best of our knowledge, there is currently no study about their influence over the results.

Additionally, besides the selected objective function used to train the model, another important choice is the color space we will work on.
Almost all colorization methods work either on a Luminance-Chrominance or on the RGB color space.
Only a few of them, such as \cite{larsson2016learning}, work on Hue-Saturation-based color spaces.
Thus, together with this chapter, another chapter called \textit{Influence of color spaces for deep learning image colorization} \cite{ballester2022influence} has been added for completeness.
It focuses on the influence of color spaces.
It also contains a more detailed review of the literature on image colorization and of the used datasets.
We refer the reader to the mentioned chapter for these reviews.

The rest of this chapter is organized as follows.
In Section~\ref{sec:sota}, we first make a review of the loss functions that have been used in the field of image colorization while connecting them with the colorization-related works.
Section~\ref{sec:framework} details the framework used to analyze the influence of the different losses, including both the chosen architecture and evaluation metrics.
Finally, in Section~\ref{sec:results} we present quantitative and qualitative colorization results on a classical image dataset, and Section~\ref{sec:archive} shows extended results on archive images.
Conclusions can be found in Section~\ref{sec:conclu}.

\section{Losses in the Colorization Literature}
\label{sec:sota}

The objective loss function summarizes the desired properties that we want the estimated outcome to satisfy. In this section, we review the losses and evaluation methods used in the literature.

Along this chapter, a color image is assumed to be defined on a bounded domain $\Omega$, a subset of $\mathbb{R}^2$. With a slight abuse of notation, we will both use the same notation to refer to the continuous setting, where $\Omega\subset\mathbb{R}^2$ is an infinite resolution image domain and $u:\Omega\to\mathbb{R}^C$, and to the discrete setting, where $\Omega$ represents a discrete domain given by a grid of $M\times N$ pixels, $M,N\in\mathbb{N}$, and $u$ is a function defined on this discrete $\Omega$ and with values in $\mathbb{R}^C$. In the latter case, $u$ is usually given by a real-valued matrix of size ${M\times N\times C}$ representing the image values. Finally, $C$ can be either equal to $3$ if $u$ is a color image, or equal to $2$ if the goal is to reconstruct the two chrominance channels and, thus, the input grayscale image is not modified during colorization.

\subsection{Error-based Losses}

In the following, the different losses used in the literature of image colorization are described and related to some representative works that capitalize on them. Table~\ref{table:losses} summarizes it.

\noindent
\textbf{MSE or squared L2 loss.}
Given two functions $u$ and $v$ defined on $\Omega$ and with values in $\mathbb{R}^C$, $C\in\mathbb{N}$, the so-called Mean Square Error (MSE) between $u$ and $v$ is defined as the squared L2 loss of their difference. That is,
\begin{equation}
\mathrm{MSE}(u,v)=\|u-v\|_{L^2(\Omega;\mathbb{R}^C)}^{2}=\int_{\Omega}\|u(x)-v(x)\|_2^2 dx,
\end{equation}
where $\|\cdot\|_2$ denotes the Euclidean norm in $\mathbb{R}^C$. In the discrete setting, it is equal to the sum of the square differences between the image values, that is
\begin{equation}
\mathrm{MSE}(u,v)=\sum_{i=1}^M\sum_{j=1}^{N}\sum_{k=1}^C (u_{i,j,k}-v_{i,j,k})^2.
\end{equation}
It has been extensively used for image colorization methods~\cite{Cheng2015,larsson2016learning,Zhang2016,Iizuka2016,isola2017image,nazeri2018image,vitoria2020chromagan} (see also Table~\ref{table:losses}), where $C=3$ if $u$ and $v$ are color images (usually the predicted and the ground truth data) or $C=2$ in the case that $u$ and $v$ are chrominance images.
Although while the training with this loss can lead to a more stable solution, it is not robust to outliers in the data and penalizes large errors while being more tolerant to small errors.

\noindent
\textbf{MAE or L1 loss with $l^1$-coupling.}
The Mean Absolute Error is defined as the L1 loss with $l^1$-coupling, that is,
\begin{equation}\label{eq:mae}
\mathrm{MAE}(u,v)=\int_{\Omega}\|u(x)-v(x)\|_{l^1} dx=\int_{\Omega}\sum_{k=1}^C |u_k(x)-v_k(x)| dx.
\end{equation}
In the discrete setting, it coincides with the sum of the absolute differences $|u_{i,j,k}-v_{i,j,k}|$. Some authors use a $l^2$-coupled version of it,
\begin{equation}
\mathrm{MAE}^c(u,v)=\sum_{i=1}^M\sum_{j=1}^{N} \sqrt{\sum_{k=1}^C (u_{i,j,k}-v_{i,j,k})^2}.
\end{equation}
Both MAE and $\mathrm{MAE}^c$ losses are robust to outliers.

To ease the non-differentiability issue in the minimization of the MAE and $\mathrm{MAE}^c$, some authors use the {\textbf{Smooth L1} or \textbf{Huber loss}}.
It is simply defined by substituting the absolute value $|\cdot |$ in \eqref{eq:mae} by
\begin{equation}
  l_H(g)=  \begin{cases} \frac{1}{2} g^2 \qquad\qquad\, \mathrm{if } \; |g|\leq\delta\\
  \delta (|g|-\frac{1}{2} \delta) \quad \mathrm{otherwise}
  \end{cases}
\end{equation}
for $g\in\mathbb{R}$.
Several works \cite{su2020instance,cao2017unsupervised,yoo2019coloring,zhang2017real,he2018deep,guadarrama2017pixcolor} use MAE, $\mathrm{MAE}^c$ or Smooth L1 losses either alone or combined with other losses (cf. Table~\ref{table:losses}).

Previous error-based losses aim to find a solution close to the ground truth.
This is counterproductive to the idea that image colorization has multiple possible solutions.
Additionally, both metrics are poorly related to perceptual quality.
Nonetheless, both metrics are the most used ones to train deep learning approaches.
In Section~\ref{sec:results} we present some numerical results together with a comparison with other kinds of losses.

Aiming at favoring a solution keeping from the ground truth not the exact values but more perceptual or style features, the following error losses have been proposed and used for colorization purposes.

\noindent
\textbf{Feature Loss.} The feature reconstruction loss \cite{gatys2015neural,johnson2016perceptual} is a perceptual loss that encourages images to have similar feature representations as the ones computed by a pretrained network, denoted here by $\Phi$.
Let $\Phi_l({u})$ be the activation of the $l$-th layer of the network $\Phi$ when processing the image ${u}$; if $l$ is a convolutional layer, then $\Phi_l({u})$ will be a feature map of size $C_l \times W_l \times H_l$.
The \emph{feature reconstruction} loss is the normalised squared Euclidean distance between feature representations, that is,
\begin{equation}
\mathcal{L}_{\text{feat}}^l(u,v) = \frac{1}{C_l W_l H_l}\left\lVert \Phi_{l}(u) - \Phi_{l}(v) \right\rVert_2^2 .
\label{eq:featureReconstruction}
\end{equation}
It penalizes the output reconstructed image when it deviates in feature content from the target.

In our experimental analysis in Section~\ref{sec:results}, we analyze the influence of the perceptual loss given by the VGG-based LPIPS~\eqref{eq:LPIPSeq}, which was introduced in~\cite{ding2021comparison} as a generalization of the perceptual loss above~\cite{johnson2016perceptual}.

\subsection{Generative Adversarial Network-based Losses}

Aiming to favor more diverse and perceptually plausible colorization results, losses based on \emph{Generative Adversarial Networks} (GANs)~\cite{goodfellow2014generative} have been introduced in the colorization literature~\cite{isola2017image,cao2017unsupervised,nazeri2018image,yoo2019coloring,vitoria2020chromagan}.
GANs are a kind of generative methods where the goal is to learn the probability distribution of the considered dataset by learning to generate new samples as if they where coming from that dataset.
In the case of GANs, the learning is done by an adversarial learning strategy.

\noindent
\textbf{Vanilla GAN.} The first GAN proposal by~\cite{goodfellow2014generative} is based on a game theory scenario between two networks competing one against another.
The first network called generator, denoted by $G$, aims to generate samples of data as similar as possible to the ones of real data $\mathcal{P}_r$.
The second network, called discriminator, aims to classify between real and generated data.
To do so, the discriminator, denoted here by $D$, is trained to maximize the probability of correctly distinguishing between real examples and samples created by the generator.
On the other hand, $G$ is trained to fool the discriminator by generating realistic examples.
The adversarial loss of the vanilla GAN is defined as: 
\begin{equation}
\mathcal{L}_{\text{adv}}(G_\theta,D_\phi)
= {\mathbb{E}}_{u\sim \mathcal{P}_r}[\log D_\phi(u)] + {\mathbb{E}}_{v\sim\mathcal{P}_{G_\theta} }[ \log (1-D_\phi(v))],
\label{eq:ganloss}
\end{equation}
and the min-max adversarial optimization problem is
\begin{equation}
\min_{G_\theta} \max_{D_\phi} \mathcal{L}_{\text{adv}}(G_\theta,D_\phi).
\label{eq:ganpb}
\end{equation}

\noindent
\textbf{Wasserstein GAN.} Although vanilla GANs have achieved good results in many domains, they have some drawbacks like convergence, vanishing gradients and mode collapse problems.
Therefore, some modifications from the original GAN have been proposed.
For example, the \emph{Wasserstein GAN} (WGAN), proposed by \cite{arjovsky2017wasserstein}, replaces the underlying Jensen--Shannon divergence from the original proposal with the Wasserstein$-1$ distance (or Earth Mover distance) between two probability distributions.
Then, the WGAN loss, $\mathcal{L}_{\text{adv,wgan}}$, and WGAN optimization problem can be defined as:
\begin{equation}
\min_{G_\theta} \max_{D_\phi\in{\cal{D}}}\mathcal{L}_{\text{adv,wgan}}(G_\theta,D_\phi) = \min_{G_\theta} \max_{D_\phi\in{\cal{D}}} \left( {\mathbb{E}}_{u\sim\mathcal{P}_r}[ D_\phi(u)] -{\mathbb{E}}_{v\sim\mathcal{P}_{G_\theta} }[D_\phi(v)] \right)
\label{eq:wganloss}
\end{equation}
where ${\cal{D}}$ denotes the set of 1-Lipschitz functions.
To enforce the $1$-Lipschitz condition, in \cite{gulrajani2017improved} the authors propose a \emph{Gradient Penalty} (GP) term constraining the L2 norm of the gradient while optimizing the original WGAN during training.
The resulting loss for the WGAN-GP can be defined as:
\begin{equation}
  \min_{G_\theta} \max_{D_\phi} \left(
  {\mathbb{E}}_{u\sim\mathcal{P}_r}[ D_\phi(u)]  -{\mathbb{E}}_{v\sim\mathcal{P}_{G_\theta}}[D_\phi(v)]
  - \lambda  {\mathbb{E}}_{\widehat{u}\sim\mathcal{\widehat{P}}}[(\| {\nabla}_{\widehat{{u}}}D(\widehat{{u}}) \|_2 -1)^2] \right)
\label{eq:wganlossgp}
\end{equation}
where $\widehat{{u}}$ is a sample defined as
\[
\widehat{{u}}  = t u  + (1-t) v,
\]
with $t$ uniformly sampled in $[0,1]$, and $u\sim\mathcal{P}_r, v\sim\mathcal{P}_{G_\theta}$.
The last term in \eqref{eq:wganlossgp} provides a tractable approximation to enforce the norm of the gradient of $D$ to be less than $1$.
The authors of \cite{gulrajani2017improved} motivated it by a theoretical result showing that the optimal discriminator $D$ contains straight lines connecting samples in the ground truth space and samples in the space of generated data.
Moreover, they experimentally observed that this technique exhibits good performance in practice.
Finally, let us observe that the minus before the gradient penalty term in \eqref{eq:wganlossgp} corresponds to the fact that the WGAN min-max objective \eqref{eq:wganlossgp} implies maximization with respect to the discriminator parameters.

In our experimental results in Section~\ref{sec:results}, we will present a comparison of several losses and we will include a combination of WGAN loss and a VGG-based LPIPS loss.
To the best of our knowledge, it has not been proposed yet.

\subsection{Distribution-based Losses}

As mentioned in Section~\ref{sec:intro}, some authors colorize an image after learning a certain probability distribution such as, for instance, a color probability distribution \cite{larsson2016learning,Zhang2016,zhang2017real,royer2017probabilistic}, or a distribution of semantic classes \cite{vitoria2020chromagan}, or directly using it for classification purposes \cite{Iizuka2016}.
The remaining of this section describes the corresponding measures of the difference between two probability distributions that have been used in the mentioned related work (see also Table~\ref{table:losses}).\\

\noindent
\textbf{Kullback--Leibler loss.}
The \emph{Kullback--Leibler} (KL) loss is the directed divergence between two probability densities $\rho$ and $\widehat{\rho}$ defined in the same space $\mathcal{Y}$. It is defined as the relative entropy from $\widehat{\rho}$ to $\rho$ which, for discrete probability densities, is given by
\begin{equation}
KL(\rho ||\widehat{\rho}) = \sum_{y\in\mathcal{Y}} \rho(y) \log{\frac{\rho(y)}{\widehat{\rho}(y)}} .
\label{eq:KL}
\end{equation}
Here, $\rho$ is usually taken as the ground truth density (sometimes as a Dirac delta or a one-hot vector on the ground truth value, or a regularized one) and $\widehat{\rho}$ the predicted one.

Some works predict a color distribution density per pixel where the color bins are associated to a fixed 2D grid in a chrominance space (\emph{e.g.},~CIE Lab in \cite{Zhang2016}). In \cite{Zhang2016}, the final color of each pixel in the inferred color image is given by the expectation (sum over the color bin centroids weighted by the histogram).
Others, such as \cite{larsson2016learning}, learn Hue-Saturation-based color distributions.
More precisely, \cite{larsson2016learning} learn the marginal distributions $\widehat{\rho}^{\mathrm{Hue}}$ and $\widehat{\rho}^{\mathrm{Chroma}}$ of Hue and Chroma, per pixel, where Chroma is related to Saturation by the formula Saturation=$\frac{\mathrm{Chroma}}{\mathrm{Value}}$ and Value=Luminance$+\frac{\mathrm{Chroma}}{2}$.
They use the KL divergence to measure the deviation between the estimated distributions and the ground truth ones.
The marginal ground truth distributions, $\rho^{\mathrm{Chroma}},\rho^{\mathrm{Hue}}$, are again defined as either a one-hot vector on the bin associated to the ground truth color, or regularized version of it.
Then, their loss is
\begin{equation}
\mathcal{L}_{}(\rho ||\widehat{\rho})=KL(\rho^{\mathrm{Chroma}} ||\widehat{\rho}^{\mathrm{Chroma}}) + \lambda c KL(\rho^{\mathrm{Hue}} ||\widehat{\rho}^{\mathrm{Hue}})
\end{equation}
where $c\in [0,1]$ is the ground truth Chroma of the considered pixel, and $\lambda=5$ in \cite{larsson2016learning}.
The authors introduce this weight depending on the Chroma multiplying the KL term on $\rho^{\mathrm{Hue}}$ to avoid Hue instability issues when Chroma approaches zero.
For inference and to sample a color value per pixel from the estimated marginal distributions, they experimentally tested that a median-based selection (a periodically modified version in the case of Hue) gives the best results.

Besides, the work \cite{vitoria2020chromagan}, uses the KL loss \eqref{eq:KL} to learn, for each image, the distribution density of semantic classes, for a fixed number of classes.
It provides information about the semantic content and objects present in the image.
In particular, they define the ground truth probability density $\rho$ of semantic classes to be the output distribution of a pre-trained VGG-16 model applied to the grayscale image, and $\widehat{\rho}$ the estimated class distribution density.

\noindent
\textbf{Cross-Entropy Loss.} Cross-Entropy loss is used for classification problems and it is sometimes referred to as logistic loss.
For discrete densities, it is defined as
\begin{equation}
CE(\rho,\widehat{\rho}) = -\sum_{y\in\mathcal{Y}} \rho(y)\log{\widehat{\rho}(y)},
\label{eq:CE}
\end{equation}
where, again, $\rho$ is usually taken as the ground truth density and $\widehat{\rho}$ the predicted one.
In the classification context, $\rho$ is often a one-hot vector equal to $1$ on the ground truth class, or a regularized version of it.
Let us also note, from \eqref{eq:KL} and \eqref{eq:CE}, that there is a relationship between the Kullback--Leibler and the Cross-Entropy losses given by
\begin{equation}
CE(\rho,\widehat{\rho}) = E(\rho) + KL(\rho ||\widehat{\rho}),
\end{equation}
where $E(\rho)$ denotes the entropy of $\rho$.

Cross-Entropy is used as a classification loss in~\cite{Iizuka2016} where the network is trained on a large-scale dataset.
The architecture is made of two encoding networks that learn local and global features and a decoder that learns the color image from these features.
The classification loss is used to guide the training of the global feature network from image label estimation. It is combined with a MSE loss that compares estimated color image with the ground truth. \\
In~\cite{Zhang2016, zhang2017real}, CE is applied on color distributions.
\cite{Zhang2016} treat the colorization problem as multinomial classification by learning a mapping from the input grayscale image to a probability distribution over possible discrete chrominance values.
CE compares the estimated distribution with the one of the ground truth.
\cite{zhang2017real} builds upon this framework and incorporates user interaction.
Finally, \cite{mouzon2019joint, pierre2020recent} stems from the resulting distributions from~\cite{Zhang2016} that, in a subsequent step, are incorporated in a variational approach~\cite{Pierre2015d}.\\

\noindent
\textbf{Log-likelihood Maximization for Diversity.}
Some works propose to generate several possible colorizations, for the same input graylevel image, by sampling over possible color distributions that are often learned by maximizing the log likelihood conditioned to the grayscale image~\cite{guadarrama2017pixcolor,royer2017probabilistic, kumar2021colorization}.

The work \emph{Pixcolor: Pixel recursive colorization} \cite{guadarrama2017pixcolor} colorizes an image by first learning the color distribution of images conditioned to a grayscale input.
It steams from autoregressive models \cite{van2016pixel,oord2016conditional,chen2018pixelsnail} that exploit the fact that a color probability distribution $p(u)$ can be in principle learned by choosing an order of the data variables $u=(u_1,u_2,\dots,u_n)\in{\mathcal{X}}$, associated to the color values of a discrete color image $u$ at its $n$ pixels (where $\mathcal{X}$ denotes the space of discrete color images), and exploiting the fact that the joint distribution can be decomposed as
\begin{equation}\label{eq:Bayes1}
p(u)=p(u_1,u_2,\dots,u_n)=p(u_1)\prod\limits_{i=2}^n p(u_i|u_1,\dots,u_{i-1}).
\end{equation}
As claimed by \cite{guadarrama2017pixcolor}, this ordering tends to capture dependencies between pixels to ensure that, at inference, colors will be consistently selected.
By working in the $YCbCr$ color space and by discretizing the $Cb$ and $Cr$ channels separately into 32 bins, they propose to model the conditional distribution of $u$ given the grayscale image $Y$ by
\begin{equation}\label{eq:Bayes2}
p(u^{b,r}|Y)=\prod \limits_i p(u^{r}_{i}|u^{b,r}_{1},\dots,u^{b,r}_{i-1},Y) p(u^{b}_{i}|u^{r}_{i},u^{b,r}_{1},\dots,u^{b,r}_{i-1},Y),
\end{equation}
where $u^{b}_{i}$ denotes the $Cb$ value for pixel $i$, $u^{r}_{i}$ its $Cr$ value, and $u^{b,r}_{i}$ its $(Cb,Cr)$ chrominance.
They train the model using maximum likelihood, with a Cross-Entropy loss per pixel. Afterwards, they perform high-resolution refinement to upscale the chrominance image at the dimensions of the original grayscale image.

In \cite{royer2017probabilistic} a feed forward network followed by an autoregressive network are used to predict for each pixel a probability distribution over all possible chrominances conditioned to the luminance.
They work in the $Lab$ color space.
$p(u^{a,b}|L)$ is factorized again as in \eqref{eq:Bayes1} and \eqref{eq:Bayes2} as the product of terms of the form $p(u^{a,b}_i|u^{a,b}_1,\dots,u^{a,b}_{i-1},L)$, that are learned on a set of training images $D$ by minimizing negative log-likelihood of the chrominance channels in the training data
\begin{equation}
\arg\min -\sum_{u \in D} \log{ p(u^{a,b}|L)}.
\label{eq:NLog}
\end{equation}
$L$ and $u^{a,b}$ denote the luminance and chrominance channels, respectively.
In order to speed up the learning, \cite{royer2017probabilistic} approximates each distribution $p(u^{a,b}_i|u^{a,b}_1,\dots,u^{a,b}_{i-1},L)$ with a mixture of 10 logistic distributions.

\cite{kumar2021colorization} also addresses the generation of multiple outputs for a given grayscale image, in this case using transformers.
They use a conditional autoregressive transfomer (a conditional variant of Axial Transformer particular self-attention with \cite{ho2019axial}) to first produce a low resolution colorization of the grayscale image (both spatial and color low resolution) that is then upsampled with two parallel networks for upsampling the spatial and color resolutions.
The model is trained to minimize the negative log-likelihood of the distributions that are estimated by each network. 

Several works combine distribution-based losses with error-based ones.
For instance, aiming to learn the distribution of color images conditioned to a grayscale version $p(u|L)$,~\cite{Deshpande2017} uses a VAE approach and log-likelihood maximization to learn a low-dimensional (latent variables) embedding of color images, combined with error losses on the output of the decoder that favor to keep color specificity (with a L2 loss that compares the projection of the generated color and ground truth images along a top-k principal components), colorfulness (with a loss that encourages rare colors to appear) and similar gradients to the ground truth color image (with a loss that compares the gradients of the generated images with the ones of the ground truth).
Moreover, the conditional distribution $p(z|L)$ of the latent variables given the grayscale image is assumed to be a Gaussian mixture and learned minimizing the conditional negative log likelihood.

The authors of~\cite{pucci2021collaborative} capitalize on capsule networks~\cite{sabour2017dynamic} to learn a color distribution over a set of quantized colors.
To that goal, they use a weighted Cross-Entropy loss where the weights are used to weight more rare colors, with a MSE loss on the $(a,b)$ channels.

\cite{kong2021adversarial} proposes a multitask network in an adversarial manner that uses a MSE loss on hue, saturation and lightness channels to perform colorization and a Cross-Entropy loss to learn a semantic segmentation.

Finally, it is worth mentioning that~\cite{ding2021comparison} compares different cost functions to train a deep neural network on four low-level vision tasks: denoising, blind image deblurring, single image super resolution and lossy image compression, although it is not done for image colorization.

In the following sections, we will present a comparison of the different loss functions for the colorization task.
To do so, we propose a baseline colorization network architecture (presented in the next section) and show experimental results for the different loss functions on the same dataset.

\begin{table}[t]
\centering
{\footnotesize
\hspace{-0.5cm}
\begin{tabular}{|c|c|c|c|c|c|c|c|c|c|c|c|c|c|c|c|c|c|c|c|c|c|c|c|}
\hline
&&&\multicolumn{5}{c|}{Using}&\multicolumn{3}{c|}{Histogram}&\multicolumn{2}{c|}{User}&\multicolumn{4}{c|}{Diverse}&\multicolumn{3}{c|}{Object}&\multicolumn{1}{c|}{Survey}\tabularnewline
&&&\multicolumn{5}{c|}{GANs}&\multicolumn{3}{c|}{prediction}&\multicolumn{2}{c|}{guided}&\multicolumn{4}{c|}{}
&\multicolumn{3}{c|}{aware}
&\multicolumn{1}{c|}{}\tabularnewline
\hline
&\rotatebox[origin=c]{90}{\cite{Cheng2015}} 
&\rotatebox[origin=c]{90}{\cite{Iizuka2016}} 
&\rotatebox[origin=c]{90}{\cite{vitoria2020chromagan}}
&\rotatebox[origin=c]{90}{\cite{nazeri2018image}}
&\rotatebox[origin=c]{90}{\cite{cao2017unsupervised}}
&\rotatebox[origin=c]{90}{\cite{yoo2019coloring}}
&\rotatebox[origin=c]{90}{\cite{antic2019deoldify}}
&\rotatebox[origin=c]{90}{\cite{larsson2016learning}}
&\rotatebox[origin=c]{90}{\cite{Zhang2016}}
&\rotatebox[origin=c]{90}{\cite{mouzon2019joint}}
&\rotatebox[origin=c]{90}{\cite{zhang2017real}}
&\rotatebox[origin=c]{90}{\cite{he2018deep}}
&\rotatebox[origin=c]{90}{\cite{Deshpande2017}}
&\rotatebox[origin=c]{90}{\cite{guadarrama2017pixcolor}}
&\rotatebox[origin=c]{90}{\cite{royer2017probabilistic}}
&\rotatebox[origin=c]{90}{\cite{kumar2021colorization}}
&\rotatebox[origin=c]{90}{\cite{su2020instance}}
&\rotatebox[origin=c]{90}{\cite{pucci2021collaborative}}
&\rotatebox[origin=c]{90}{\cite{kong2021adversarial}}
&\rotatebox[origin=c]{90}{winner of \cite{Gu_2019_CVPR_Workshops}}
\tabularnewline
\hline
MAE & & &  & &$\bullet$ & & & & & & & & &     $\bullet$ & & & & & & $\bullet$  \tabularnewline
\hline
smooth-L1 & & &  & & & $\bullet$ &  & & & &  $\bullet$  &$\bullet$ & & & & & $\bullet$ & & &  \tabularnewline
\hline
MSE & $\bullet$& $\bullet$&   $\bullet$& $\bullet$&   & & & $\bullet$ & $\bullet$ & $\bullet$ & & &  $\bullet$ & & & & & $\bullet$ & $\bullet$ & $\bullet$  \tabularnewline
\hline
GANs & & & $\bullet$ & $\bullet$ & $\bullet$ & $\bullet$ & $\bullet$ & & & & & & & & & & & & $\bullet$ & \tabularnewline
\hline
KL {\scriptsize on distributions}& & & & & & & &  $\bullet$ & & & & & & & & & &  & & \tabularnewline
\hline
CE {\scriptsize on distributions} & & & & & & & & &$\bullet$ & $\bullet$ & $\bullet$ & & & & & &  &$\bullet$  & &  \tabularnewline
\hline
KL {\scriptsize for classification}& & & $\bullet$  & & & & & & & & & & & & & & &  & &  \tabularnewline
\hline
CE {\scriptsize for classification } & & $\bullet$&  & & & & & & &  & & & & & & & & &$\bullet$  &    \tabularnewline
\hline
neg log-likelihood & & & & & & & & & &  &  & &  $\bullet$ &  $\bullet$ & $\bullet$ & $\bullet$ & & & &  \tabularnewline
\hline
Perceptual & & & & & & & $\bullet$ & & &  & &$\bullet$ & & & & & & & &  \tabularnewline
\hline
\end{tabular}}
\caption{Losses used to train deep learning methods for image colorization. CE stands for Cross-Entropy and KL for Kullback-Leibler divergence.}
\label{table:losses}
\end{table}

\section{Proposed Colorization Framework}
\label{sec:framework}

In this section, we present the framework used to study the influence of the chosen objective loss on the estimated images colorization results.
First we detail the architecture and secondly the dataset used for both training and testing.
Note that the same architecture and training procedure is used in the other chapter \textit{Influence of Color Spaces for Deep Learning Image Colorization} \cite{ballester2022influence}.

\subsection{Detailed Architecture}

\begin{figure}[t]
\centering
\includegraphics[width=\linewidth]{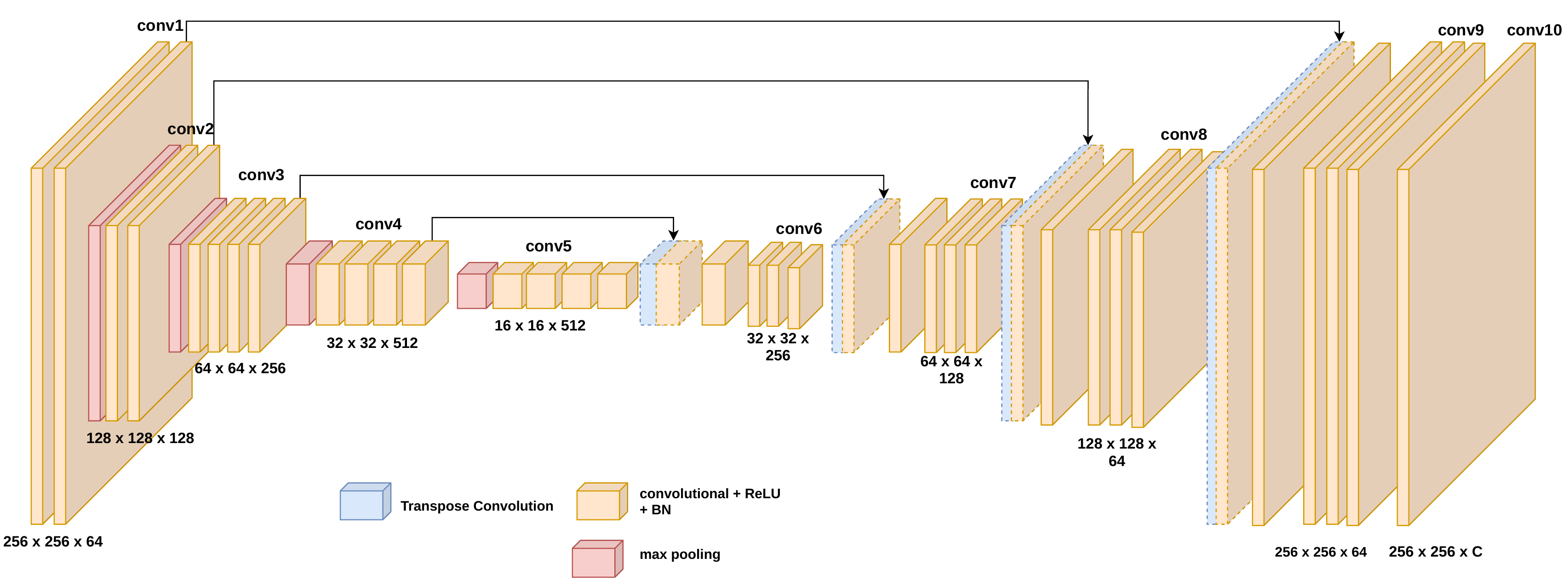}
\caption{Summary of the baseline U-Net architecture used in our experiments. It outputs a $256\times256\times C$ image, where $C$ stands for the number of channels, being equal to $2$ when estimating the missing chrominance channels and to $3$ when estimating the RGB components.}
\label{fig:unet-architecture}
\end{figure}

The architecture used in our experiments is an encoder-decoder U-Net composed of five stages. Figure~\ref{fig:unet-architecture} displays a summary of the whole architecture.
All convolutional blocks are composed of two 2D convolutional layers with kernels of kernel size equal to $3 \times 3$, each one followed by 2D batch normalization and a ReLU activation.
For the encoder, downsampling is done by using a max pooling operator after each convolutional block.
After downsampling, the number of filters is doubled in the following block.
For the decoder, upsampling is done by using 2D transpose convolutions (with $4 \times 4$ kernels with stride 2).
At a given stage, the corresponding encoder and decoder blocks are linked with skip connections: feature maps from the encoder are concatenated with the ones from the corresponding upsampling path and fused using $1 \times 1$ convolutions. More details can be found in Table~\ref{tab:arch-details}.

The encoder architecture is identical to the CNN part of a VGG network~\cite{simonyan2017very}.
It allows us to start from pretrained weights initially used for ImageNet classification.

\begin{table}[t]
\centering
\setlength{\tabcolsep}{0.4cm}
\begin{tabular}{lc}
\hline
\textbf{Layer type}           & \textbf{Output resolution} \\ \hline
Input                         & 3 x H x W                             \\
Conv1 + Max-pooling           & 64 x H/2 x W/2                        \\
Conv2 + Max-pooling           & 128 x H/4 x W/4                       \\
Conv3 + Max-pooling           & 256 x H/8 x W/8                       \\
Conv4 + Max-pooling           & 512 x H/16 x W/16                     \\
Conv5 + Conv. Transpose (I)   & 512 x H/8 x W/8                       \\
Conv6 + Conv. Transpose (II)  & 256 x H/4 x W/4                       \\
Conv7 + Conv. Transpose (III) & 128 x H/2 x W/2                       \\
Conv8 + Conv. Transpose (IV)  & 64 x H x W                            \\
Conv9                         & 64 x H x W                            \\
Conv10                        & C x H x W                             \\ \hline
\end{tabular}
\caption{Detailed architecture and output resolution for each block.}
\label{tab:arch-details}
\end{table}

The training settings are described as follows:
\begin{itemize}
    \item Optimizer: Adam
    \item Learning rate: 2e-5.
    \item Batch size: 16 images (10-11 GB RAM on Nvidia Titan V).
    \item All images are resized to $256 \times 256$ for training which enables using batches.
    In practice, to keep the aspect ratio, the image is resized such that the smallest dimension matches $256$. If the other dimension remains larger than $256$, we then apply a random crop to obtain a square image. Note that the random crop is performed using the same seed for all trainings.
\end{itemize}

More details regarding this framework are given in the other chapter \emph{Influence of Color Spaces for Deep Learning Image Colorization} \cite{ballester2022influence}.

\subsection{Quantitative Evaluation Metrics Used in Colorization Methods}

For the last twenty years, colorization methods have mostly been evaluated with MAE, MSE, Peak signal-to-noise ratio (PSNR) and Structural Similarity Index (SSIM) metrics~\cite{wang2004image}.

In the context of colorization, the PSNR measures the ratio between the maximum value of a color target image $u:\Omega\rightarrow \mathbb{R}^C$ and the mean square error (MSE) between $u$ and a colorized image $v:\Omega\rightarrow \mathbb{R}^C$ with $\Omega\in\mathbb{Z}^2$ a discrete grid of size $M\times N$. That is,
\begin{align}
\rm{PSNR}(u,v) = 20&\log_{10}(\max{u})   \nonumber \\
&- 10\log_{10} \left(\frac{1}{CMN}\sum_{k=1}^C\sum_{i=1}^M\sum_{j=1}^N (u(i,j,k)-v(i,j,k))^2\right),
\end{align}
where $C=3$ when working in the RGB color space and $C=2$ in any luminance-chrominance color space as YUV, Lab and YCbCr. The PSNR score is considered as a reconstruction measure tending to favor methods that will output results as close as possible to the ground truth image in terms of the MSE.

SSIM intends to measure the perceived change in structural information between two images. It combines three measures to compare images color~($l$), contrast~($c$) and structure~($s$):
\begin{equation}
\rm{SSIM}(u,v) = l(u,v)c(u,v)s(u,v)\\
= \frac{\left(2\mu_u\mu_v\right)+c_1}{\mu_u^2+\mu_v^2+c_1}\frac{\left(2\sigma_u\sigma_v+c_2\right)}{\sigma_u^2+\sigma_v^2+c_2}\frac{\left(\sigma_{uv}+c_3\right)}{\sigma_u\sigma_v+c_3}
\end{equation}
where $\mu_u$ (resp. $\sigma_u$) is the mean value (resp. the variance) of image $u$ values and $\sigma_{uv}$ the covariance of $u$ and $v$. $c_1$, $c_2$, $c_3$ are regularization constants that are used to stabilize the division for images with mean or standard deviation close to zero.

More recently, other perceptual metrics based on deep learning have been proposed: the Fréchet Inception Distance (FID) \cite{heusel2017gans}, and a Learned Perceptual Image Patch Similarity (LPIPS) \cite{Zhang_2018_CVPR}.
They have been widely used in image editing for their ability to correlate well with human perceptual similarity.
FID~\cite{heusel2017gans} is a quantitative measure used to evaluate the quality of the outputs' generative model and which aims at approximating human perceptual evaluation.
It is based on the Fréchet distance~\cite{dowson1982frechet} which measures the distance between two multivariate Gaussian distributions.
FID is computed between the feature-wise mean and covariance matrices of the features extracted from an Inception v3 neural network applied to the input images $(\mu_r,\Sigma_r)$ and those of the generated images $(\mu_g,\Sigma_g)$:
\begin{equation}
\rm{FID}((\mu_r,\Sigma_r),(\mu_g,\Sigma_g)) = \Vert \mu_r - \mu_g \Vert^2_2 + Tr(\Sigma_r + \Sigma_g - 2\Sigma_r \Sigma_g)^{1/2}.
\end{equation}

LPIPS~\cite{Zhang_2018_CVPR} computes a weighted L2 distance between deep features of a pair of images $u$ and $v$:
\begin{equation}\label{eq:LPIPSeq}
\rm{LPIPS}(u,v) = \sum_l\frac{1}{H_lW_l}\sum_{i=1}^{H_l}\sum_{j=1}^{W_l}\|\omega_l\odot (\Phi_l(u)_{i,j} - \Phi_l(v)_{i,j} )\|_2^2 ,
\end{equation}
where $H_l$ (resp. $W_l$) is the height (resp. the width) of feature map $\Phi_l$ at layer $l$ and $\omega_l$ are weights for each features. Note that features are unit-normalized in the channel dimension.

Other quantitative metrics can be found in literature for image colorization.
Accuracy~\cite{nazeri2018image} measures the ratio between the number of pixels that have the same color information as the source and the total number of pixels.
Raw accuracy (AuC)~\cite{Zhang2016} computes the percentage of predicted pixel colors within a threshold of the L2 distance from the ground truth in ab color space.
The result is then swept across thresholds from 0 to 150 to produce a cumulative mass function.
\cite{Deshpande2017} evaluates colorfulness as the MSE on histograms.
\cite{royer2017probabilistic} verify if the framework produces vivid colors by computing the average perceptual saturation~\cite{lubbe2010colours}.
Other works evaluate the capability of a classification network to infer the right class to the generated image~\cite{Zhang2016,he2018deep}.
\cite{Zhang2016} feeds the generated image to a classification network and observe if the classifier performs well.

Note that all models that are trained with a L2 loss will more likely get better PSNR or MSE as the L2 loss is correlated with the evaluation.

Table~\ref{table:eval} summarizes the quantitative evaluation metrics more generally used in the literature of image colorization.
In our experiments, we choose to rely on the more generally used and more recent ones, namely, L1 (MAE), L2 (MSE), PSNR, SSIM, LPIPS and FID.

\begin{table}[t]
\centering
{\footnotesize
\hspace{-0.5cm}\begin{tabular}{| c| c| c| c| c|c|c|c| c| c| c  |}
\hline
&\multicolumn{7}{c|}{\textbf{Quantitative}} & \multicolumn{3}{c|}{\textbf{User Study}}\tabularnewline
 \hline
&\rotatebox[origin=c]{90}{\textbf{L1/ MAE}} &\rotatebox[origin=c]{90}{\textbf{L2/ MSE}} & \textbf{PSNR} & \textbf{SSIM} & \textbf{LPIPS} & \textbf{FID} & \textbf{Other}& \rotatebox[origin=c]{90}{AMT Fooling Rate}& \rotatebox[origin=c]{90}{Naturalness} & Other \tabularnewline
 \hline
\cite{Cheng2015} & & & & $\bullet$ & & & & & &  \tabularnewline
\hline
\cite{Iizuka2016} & & & & & & &$\bullet$  &  &$\bullet$  &\tabularnewline
\hline
\multicolumn{11}{|c|}{Using GANS}\tabularnewline
\hline
\cite{vitoria2020chromagan} && & $\bullet$ & & & & &  & $\bullet$ & \tabularnewline
\hline
\cite{nazeri2018image} & $\bullet$& & & & & &   $\bullet$  & & & \tabularnewline
\hline
\cite{cao2017unsupervised} &  & $\bullet$ & $\bullet$ & & & &    & & &$\bullet$\tabularnewline
\hline
\cite{yoo2019coloring} &  &  &  & & $\bullet$& &  & & &$\bullet$\tabularnewline
\hline
\multicolumn{11}{|c|}{Histograms Prediction}\tabularnewline
\hline
\cite{larsson2016learning} & &$\bullet$&$\bullet$& & &&$\bullet$& &&  \tabularnewline
\hline
\cite{Zhang2016} & && &  & &  &$\bullet$ & $\bullet$ & & \tabularnewline
\hline
\multicolumn{11}{|c|}{User Guided}\tabularnewline
\hline
\cite{zhang2017real}&&&$\bullet$&&&&&$\bullet$&&\tabularnewline
\hline
\cite{he2018deep}&&&$\bullet$&&&&$\bullet$&$\bullet$&&\tabularnewline
\hline
\multicolumn{11}{|c|}{Diverse}\tabularnewline
\hline
\cite{Deshpande2017}&&$\bullet$&&&&&&&&$\bullet$\tabularnewline
\hline
\cite{guadarrama2017pixcolor}&&&&$\bullet$&&&&$\bullet$&&\tabularnewline
\hline
\cite{royer2017probabilistic}&&&&&&&&&&$\bullet$\tabularnewline
\hline
\cite{kumar2021colorization}&&&&&&$\bullet$&&$\bullet$&&\tabularnewline
\hline
\multicolumn{11}{|c|}{Object Aware}\tabularnewline
\hline
\cite{su2020instance} &  &  & $\bullet$ & $\bullet$& $\bullet$ & &  & & & \tabularnewline
\hline
\cite{pucci2021collaborative} &  &  & $\bullet$ & & $\bullet$ & &  & & & \tabularnewline
\hline
\cite{kong2021adversarial} &  &  & $\bullet$ &$\bullet$ &  & & $\bullet$ & & & \tabularnewline
\hline
\multicolumn{11}{|c|}{Survey}\tabularnewline
\hline
\cite{Gu_2019_CVPR_Workshops} &&&$\bullet$&$\bullet$&&& & &&$\bullet$\tabularnewline
\hline
\end{tabular}}
\caption{Evaluation metrics used by deep learning methods for image colorization.}\label{table:eval}
\end{table}

\section{Experimental Analysis}
\label{sec:results}

To compare the influence of the objective loss in the resulting colorization results, we train the network described in Section~\ref{sec:framework} by changing the objective loss.
In particular, we train the network with the L1 loss, the L2 loss, the VGG-based LPIPS, the combination of WGAN plus L2 losses, and the combination of WGAN and VGG-based LPIPS.
To the best of our knowledge, the combination of the VGG-based LPIPS loss with a WGAN training procedure is novel and has not been proposed in the recent literature.

For each of the these losses, depending on the chosen color space, we estimate:
\begin{itemize}
    \item either the two $(a,b)$ chrominance channels given the luminance channel $L$ as input;
    \item or the three $(R,G,B)$ color channels given a grayscale image as input.
\end{itemize}

In this section, we present a quantitative and qualitative comparison for all of these combinations.
Note that to compute the VGG-based LPIPS loss, the output colorization always has to be converted to RGB (in a differentiable way), even for Lab color space, because this loss is computed with a pre-trained VGG expecting RGB images as input.
To this end, we have used the Kornia implementation of differentiable color space conversions~\cite{riba2020kornia}.

Throughout our experiments we use the COCO dataset~\cite{lin2014microsoft}, containing various natural images of different sizes.
COCO is divided into three sets that approximately contain 118k, 5k and 40k images that, respectively, correspond to the training, validation and test sets.
Note that we carefully remove all grayscale images, which represents around $3\%$ of the overall amount of each set.
Although larger datasets such as ImageNet have been regularly used in the literature, COCO offers a sufficient number and a good variety of images so we can efficiently train and compare numerous models.
While the training is done on batches of square $256\times 256$ images, for testing we apply the network to images at their original resolution.

\subsection{Quantitative Evaluation}

\begin{table}[t]
\centering
\setlength{\tabcolsep}{0.15cm}
\renewcommand{\arraystretch}{1.2}
\begin{tabular}{c|c|cccccc}
\hline
\textbf{Color space} & \textbf{Loss function} & \textbf{MAE} $\downarrow$ & \textbf{MSE} $\downarrow$ & \textbf{PSNR}  $\uparrow$ & \textbf{SSIM} $\uparrow$ & \textbf{LPIPS} $\downarrow$ & \textbf{FID} $\downarrow$ \\
\hline
Lab & L1         & 0.04407 & 0.00589 & 22.3020 & \textbf{0.9268} & 0.1587 & 8.8109  \\
Lab & L2         & 0.04488 & 0.00585 & 22.3283 & \underline{0.9250} & 0.1613 & 8.1517 \\
Lab & LPIPS      & \textbf{0.04374} & \textbf{0.00566} & \textbf{22.4699} & 0.9228 & \textbf{0.1403} & \underline{3.2221} \\
Lab & WGAN+L2    & 0.04459 & 0.00582 & 22.3512 & 0.9243 & 0.1609 & 7.6127  \\
Lab & WGAN+LPIPS & \underline{0.04383} & \underline{0.00568} & \underline{22.4541} & 0.9223 & \underline{0.1406} & \textbf{3.1045} \\
\hline
\hline
RGB & L1         & \textbf{0.04385} & 0.00587 & 22.3119 & \textbf{0.9268} & 0.1583 & 8.0125 \\
RGB & L2         & \underline{0.04458} & \underline{0.00587} & \underline{22.3136} & \underline{0.9255} & 0.1606 & 7.4223 \\
RGB & LPIPS      & 0.04573 & \textbf{0.00577} & \textbf{22.3892} & 0.9196 & \textbf{0.1429} & \underline{3.0576} \\
RGB & WGAN+L2    & 0.05256 & 0.00651 & 21.8667 & 0.8559 & 0.2469 & 15.4780 \\
RGB & WGAN+LPIPS & 0.04901 & 0.00679 & 21.6806 & 0.9137 & \underline{0.1495} & \textbf{2.6719} \\
\hline
\end{tabular}
\caption{Quantitative evaluation of colorization results for different loss functions. Metrics are used to compare ground truth to every images in the 40k test set. Best and second best results by column are in bold and underlined respectively.}
\label{table:Quantitativeresults}
\end{table}

Table~\ref{table:Quantitativeresults} shows the quantitative results comparing five losses, namely, the L1 loss, the L2 loss, the  VGG-based LPIPS, the combination of WGAN plus L2 losses, and the combination of WGAN and VGG-based LPIPS (denoted in Table~\ref{table:Quantitativeresults} as L1, L2, LPIPS, WGAN+L2, and WGAN+LPIPS, respectively).
The first five rows display this assessment when the used color space is Lab (\emph{i.e.},~the model estimates the two ab chrominance channels), while for the last five rows the used color space is RGB  (\emph{i.e.},~the model estimates the three RGB color channels).
In particular, let us remark that the quantitative evaluations are always performed in the final RGB color space.
Thus, even when the model is trained to estimate the ab chrominance channels, the resulting Lab color image is converted to the RGB color space to compute the evaluation metrics.

From the results in Table~\ref{table:Quantitativeresults} we observe that for the analyzed dataset, the models trained with the VGG-based LPIPS loss function provide overall better quantitative results, for both Lab and RGB color spaces.
This is especially true for the perceptual metrics LPIPS and FID, as they are strongly correlated to this loss function.
The fact that the VGG-based LPIPS training loss is computed on RGB color space (as this loss is computed with a pre-trained VGG expecting RGB images as input) and also are all quantitative results might be related to the performance (see also \cite{ballester2022influence}).
In the same spirit, we can observe a slight correlation between the used training loss and the quantitative metric.
For instance, when training with L1, MAE results are better.
However, we can see that L2 loss is not at the top in any of the metrics, while we could have expected in the case of MSE or PSNR, but this is not the case.

Nevertheless, no strong tendency clearly emerges from this table: for many metrics, the different losses do not differ so much from one another and could be in the margin of error.
From our analysis, we hypothesize that, apart from the chosen objective function, the network architecture design, and the training process, may play a very important role as a prior on the colorization operator.
Further analysis will be done on that matter.

Finally, let us mention the importance of user-based quantitative studies to properly assess colorizations results.
It is not just important in cases where no ground truth is available, such as the ones of old archive photographs, but also due to the fact that multiple colorizations are always possible.
Several works propose different user-based metrics such as, \emph{e.g.},~\emph{naturalness} or \emph{fooling rate}.
Nevertheless, efforts should be made on a widely accepted protocol and a widespread user study metric.

\subsection{Qualitative Evaluation}

\begin{figure}[t]
\begin{center}
\begin{tabular}{ccccc}
    \includegraphics[width=0.2\linewidth]{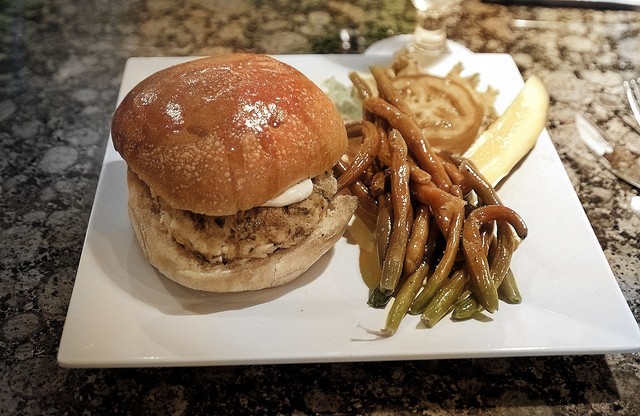}
& 
    \includegraphics[width=0.2\linewidth]{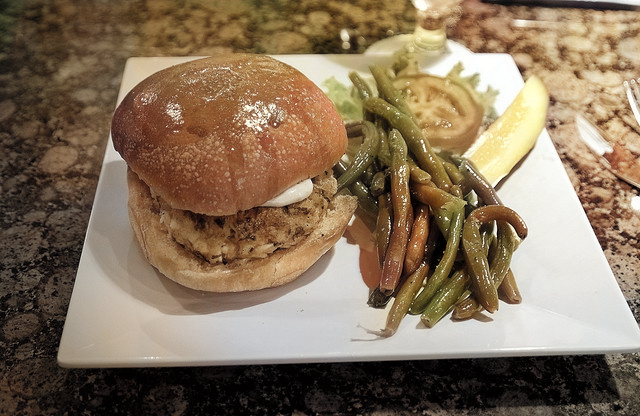}
& 
    \includegraphics[width=0.2\linewidth]{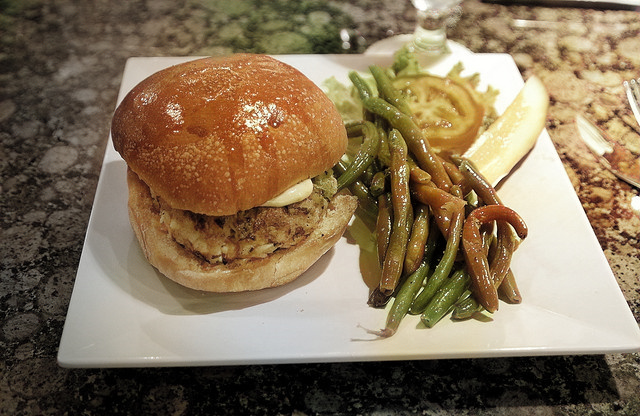}
& 
    \includegraphics[width=0.2\linewidth]{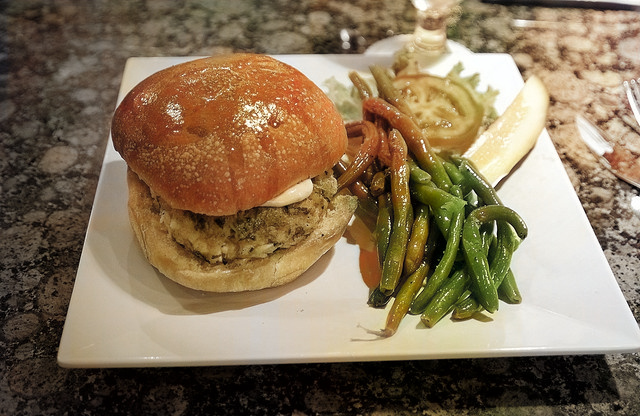}
&
    \includegraphics[width=0.2\linewidth]{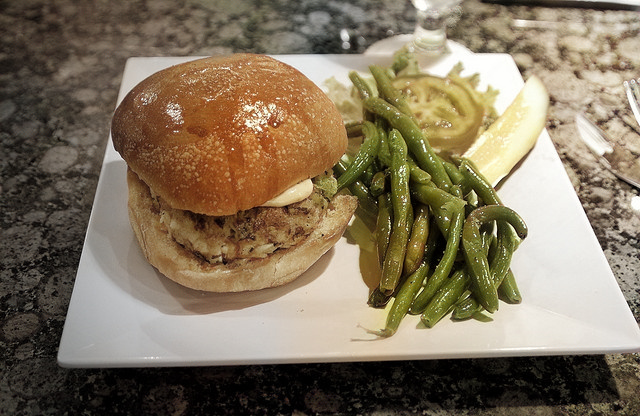}
\\
    \includegraphics[width=0.2\linewidth]{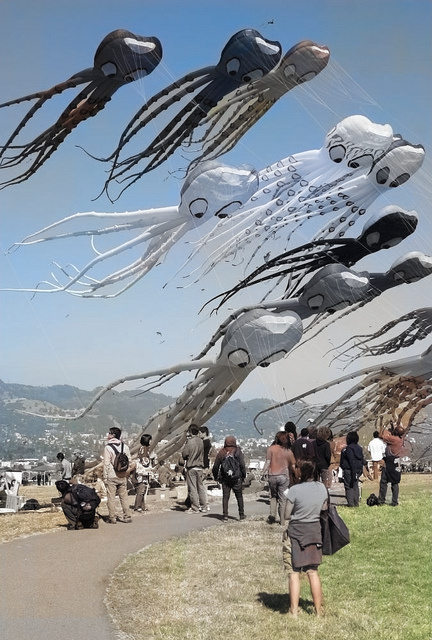}
& 
    \includegraphics[width=0.2\linewidth]{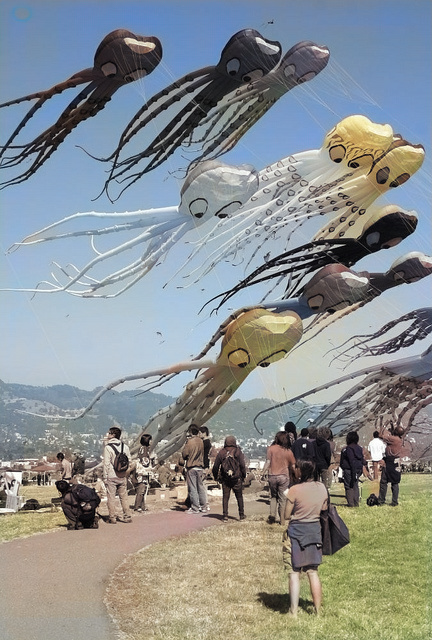}
& 
    \includegraphics[width=0.2\linewidth]{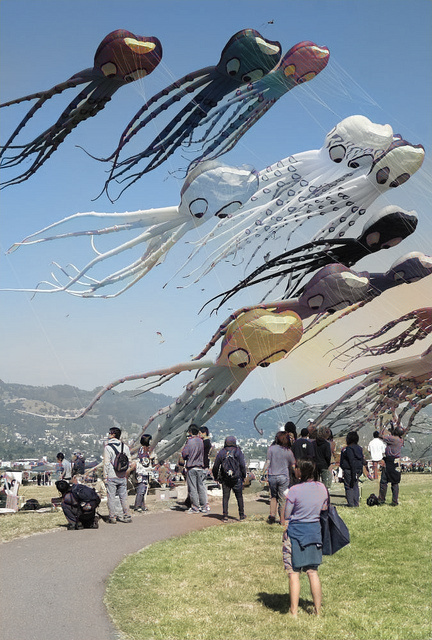}
& 
    \includegraphics[width=0.2\linewidth]{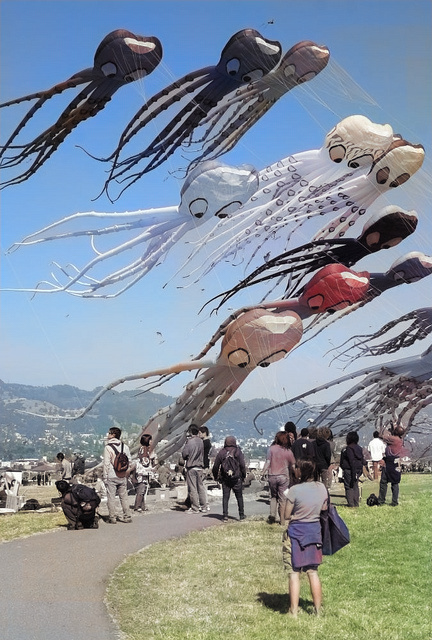}
&
\includegraphics[width=0.2\linewidth]{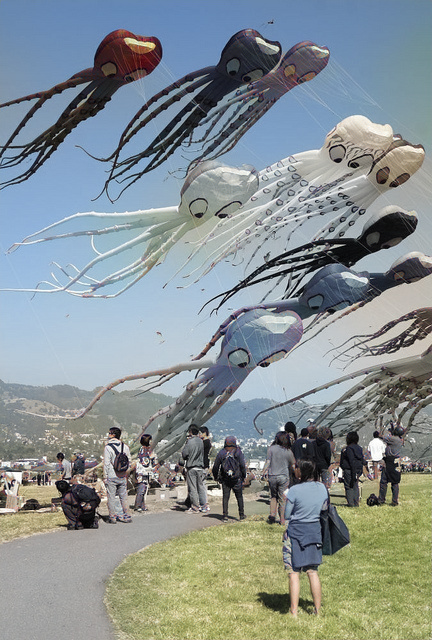}
\\
    \includegraphics[width=0.2\linewidth]{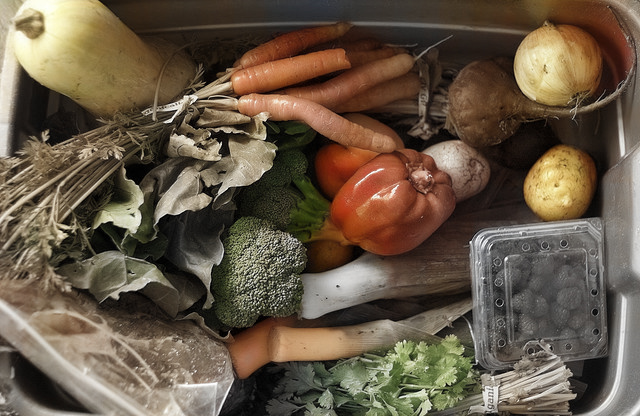}
& 
    \includegraphics[width=0.2\linewidth]{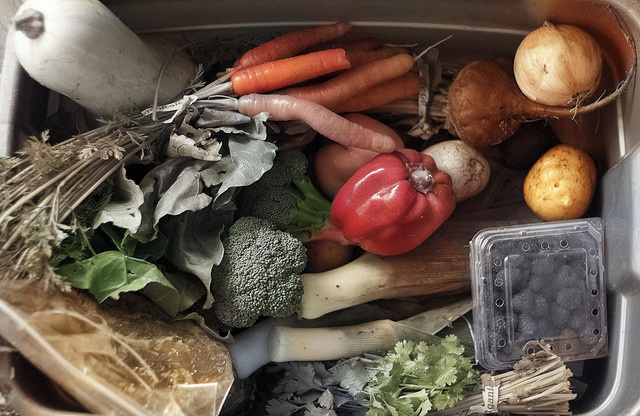}
& 
    \includegraphics[width=0.2\linewidth]{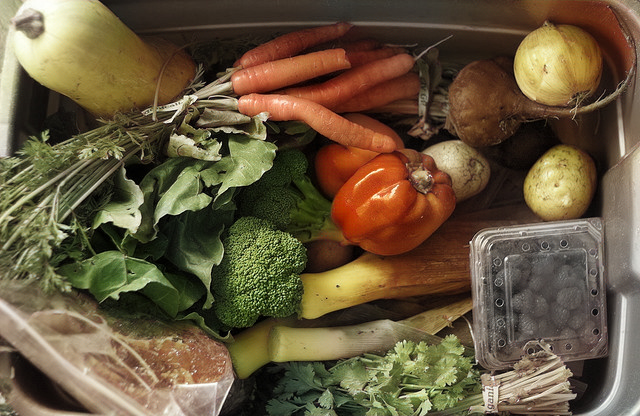}
& 
    \includegraphics[width=0.2\linewidth]{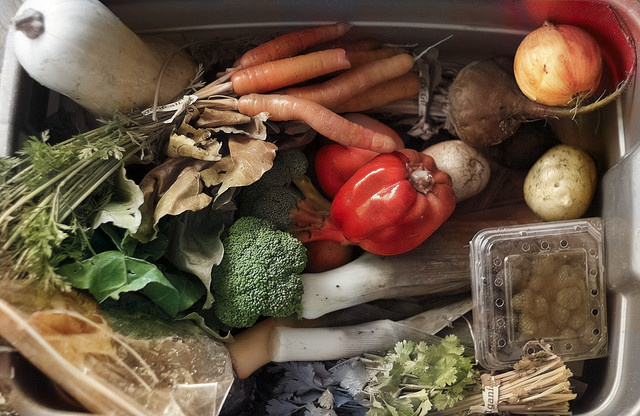}
&
    \includegraphics[width=0.2\linewidth]{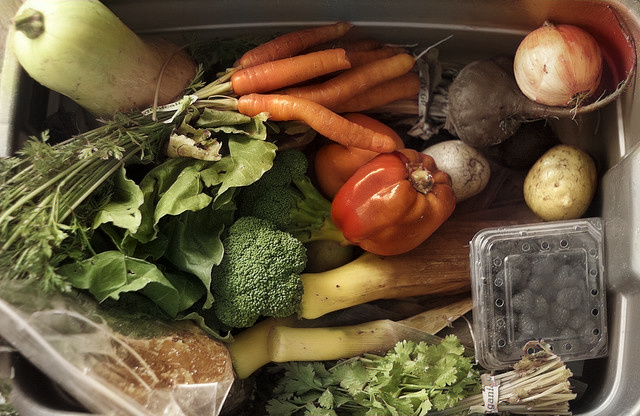}
\\
    \includegraphics[width=0.2\linewidth]{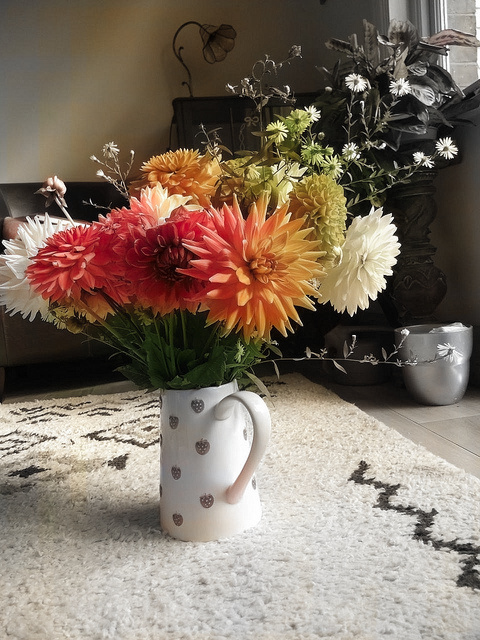}
& 
    \includegraphics[width=0.2\linewidth]{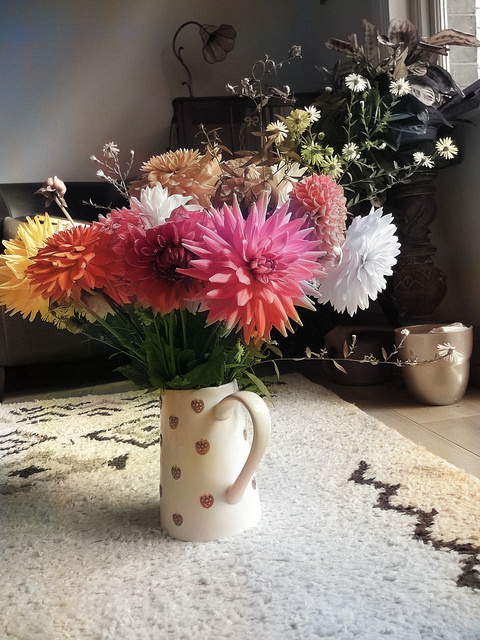}
&
    \includegraphics[width=0.2\linewidth]{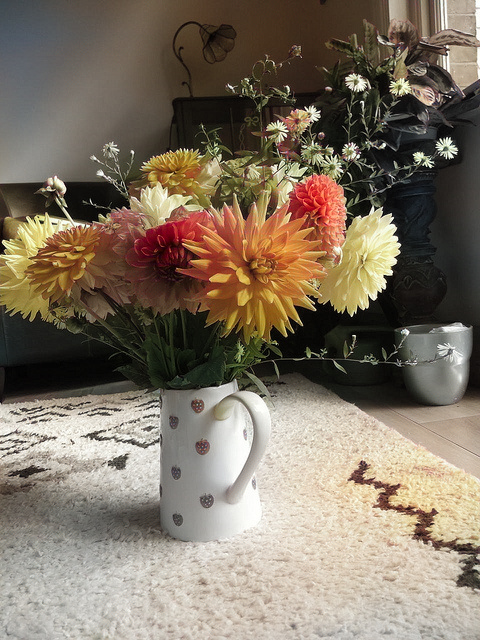}
& 
    \includegraphics[width=0.2\linewidth]{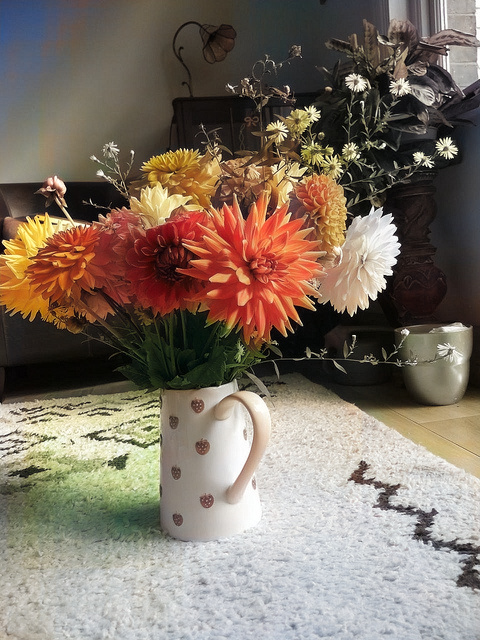}
& 
    \includegraphics[width=0.2\linewidth]{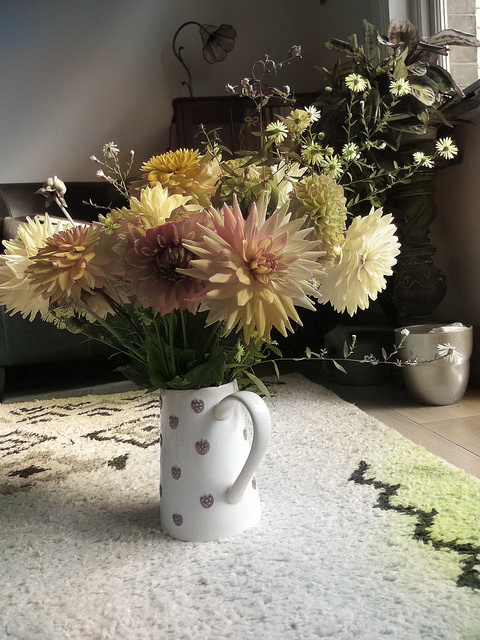}
\\
Lab-L1  & 
Lab-L2  & 
Lab-LPIPS  & 
Lab-WGAN+L2 &
Lab-WGAN+LPIPS
\\
\end{tabular}
\caption{Examples where multiple objects are in the same image. Five losses are compared, namely, L1, L2, LPIPS, WGAN+L2 and WGAN+LPIPS. The used color space is Lab for all the cases (that is, the model estimates two ab chrominance channels).}
\label{manyobjectsLab}
\end{center}
\end{figure}

\begin{figure}[t] \begin{center}
\begin{tabular}{ccccc}
    \includegraphics[width=0.2\textwidth]{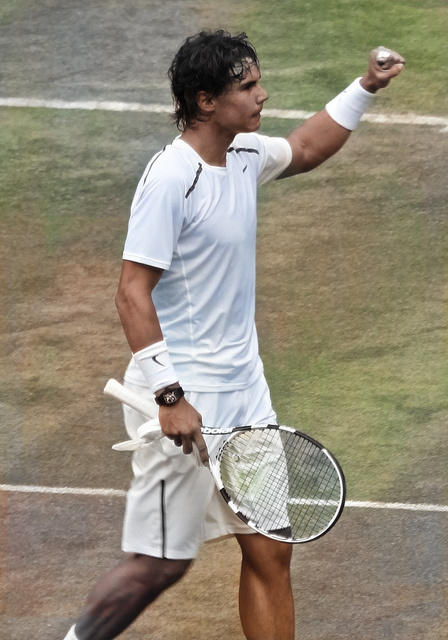}
& 
    \includegraphics[width=0.2\textwidth]{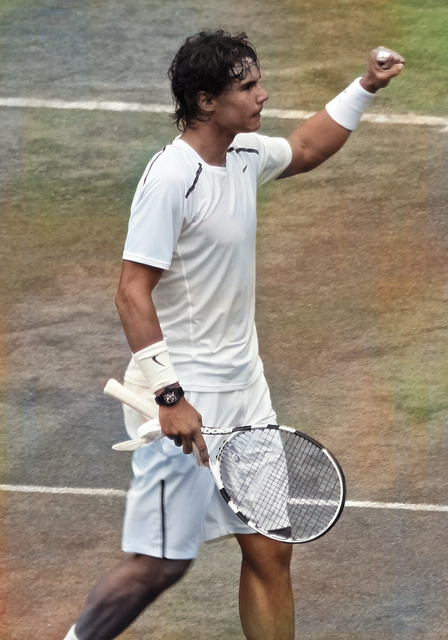}
    & 
    \includegraphics[width=0.2\textwidth]{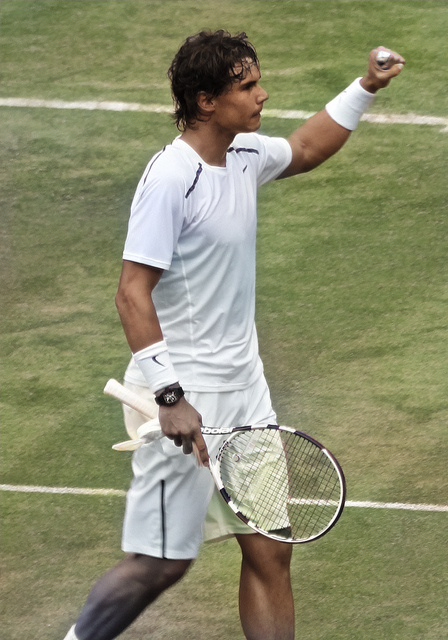}
& 
    \includegraphics[width=0.2\textwidth]{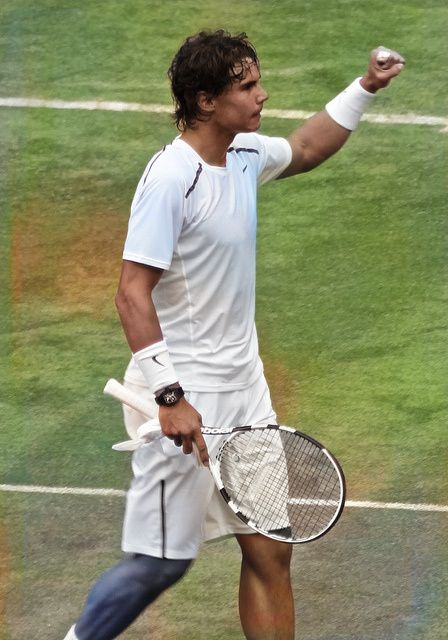}
        & 
    \includegraphics[width=0.2\textwidth]{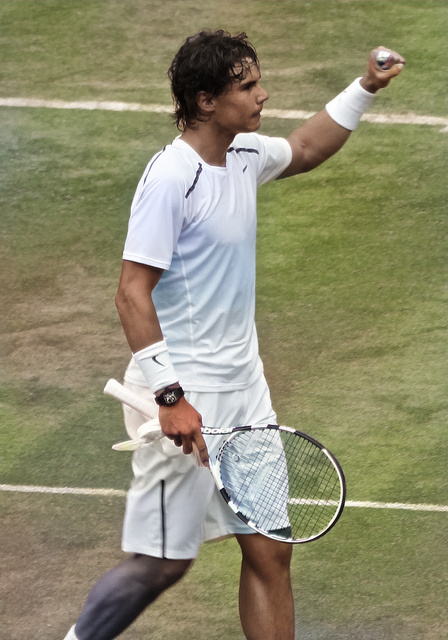}
\\
    \includegraphics[width=0.2\textwidth]{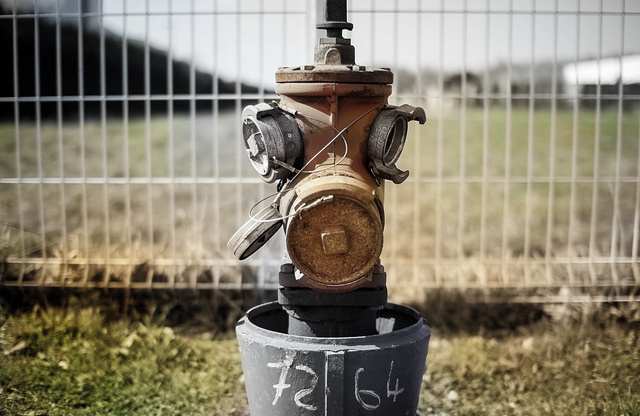}
& 
    \includegraphics[width=0.2\textwidth]{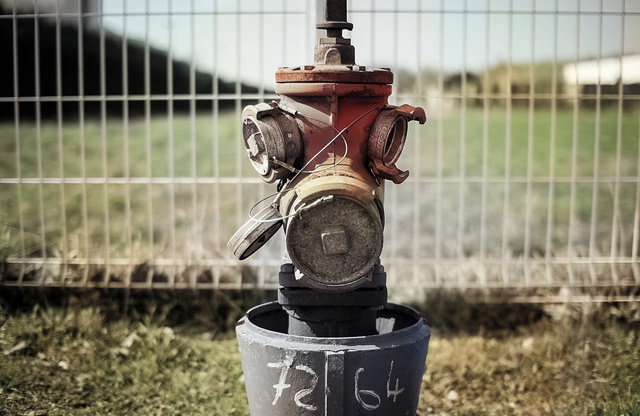}
    & 
    \includegraphics[width=0.2\textwidth]{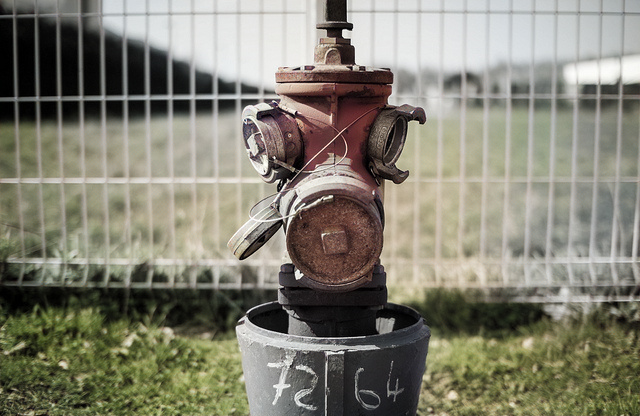}
& 
    \includegraphics[width=0.2\textwidth]{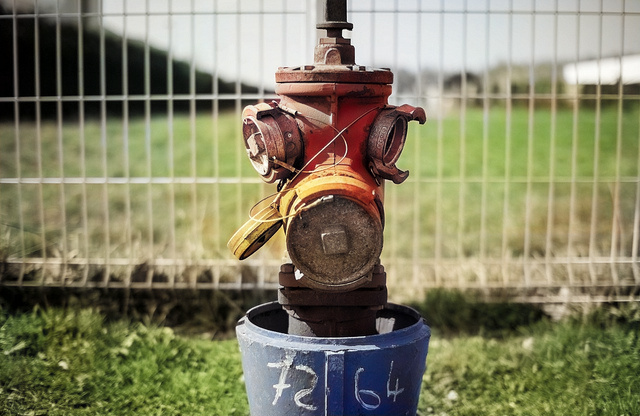}
        & 
    \includegraphics[width=0.2\textwidth]{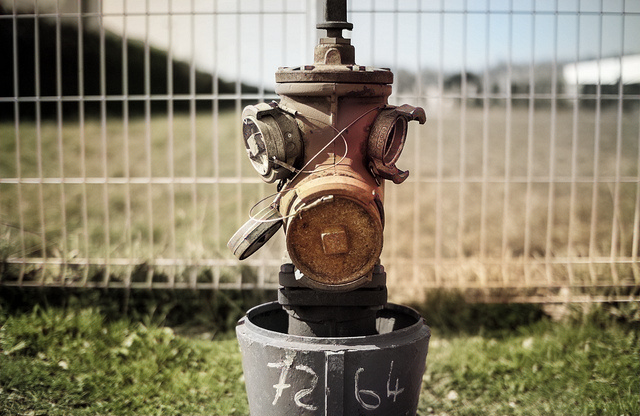}
\\
    \includegraphics[width=0.2\textwidth]{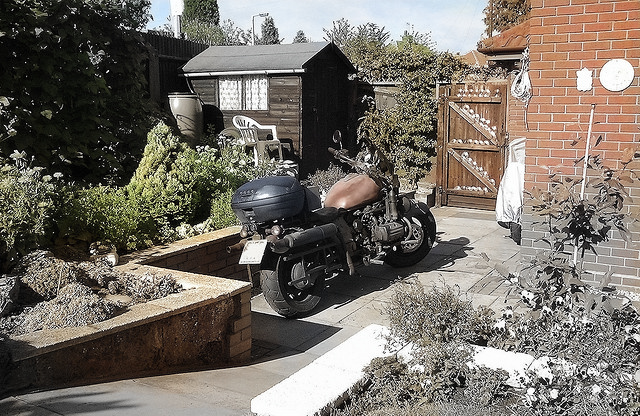}
& 
    \includegraphics[width=0.2\textwidth]{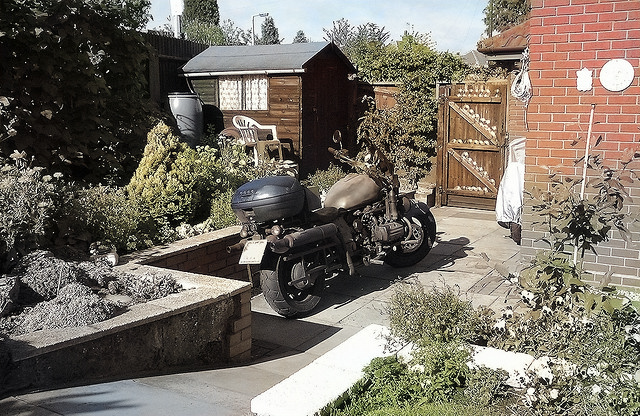}
    & 
    \includegraphics[width=0.2\textwidth]{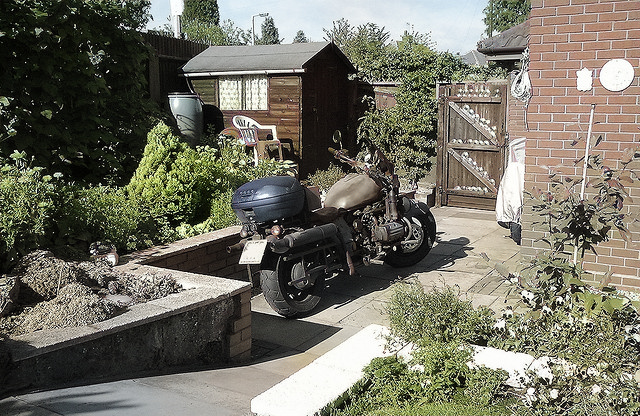}
& 
    \includegraphics[width=0.2\textwidth]{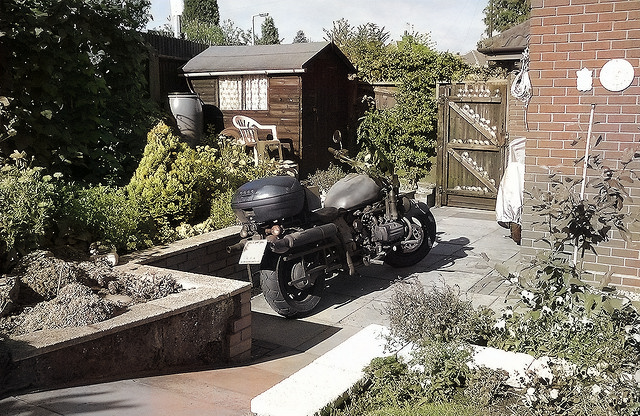}
        & 
    \includegraphics[width=0.2\textwidth]{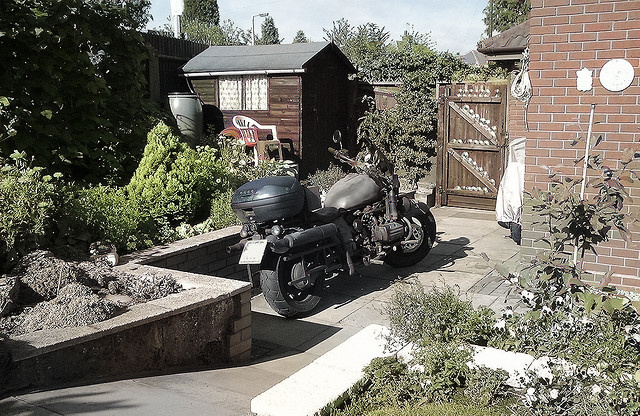}
\\
 Lab-L1  & 
 Lab-L2  & 
 Lab-LPIPS  & 
Lab-WGAN+L2 &
Lab-WGAN+LPIPS
\\
 \end{tabular}
    \caption{Examples to evaluate shyniness of the results. Five losses are compared, namely, L1, L2, LPIPS, WGAN+L2 and WGAN+LPIPS. The used color space is Lab for all the cases.}\label{shinyLab}
 \end{center} \end{figure}

\begin{figure}[t] \begin{center}
\begin{tabular}{ccccc}
    \includegraphics[width=0.2\textwidth]{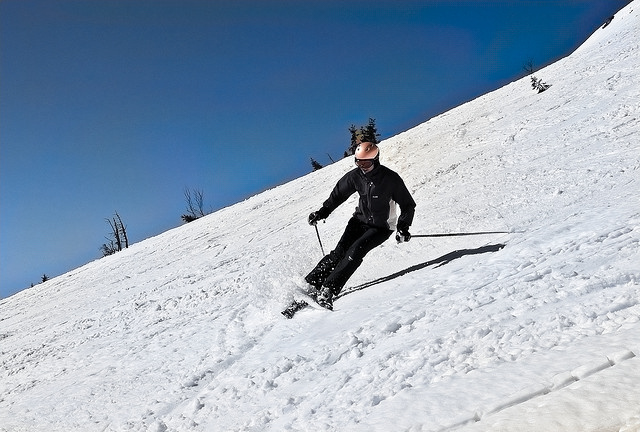}
& 
    \includegraphics[width=0.2\textwidth]{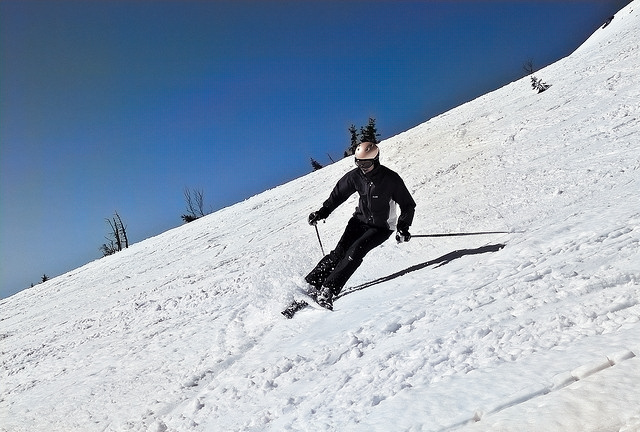}
    & 
    \includegraphics[width=0.2\textwidth]{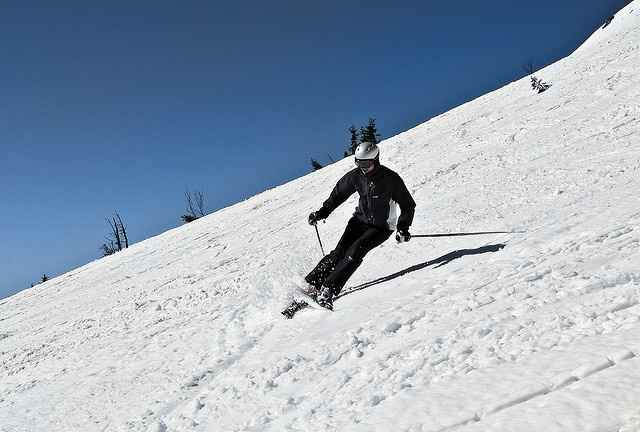}
& 
    \includegraphics[width=0.2\textwidth]{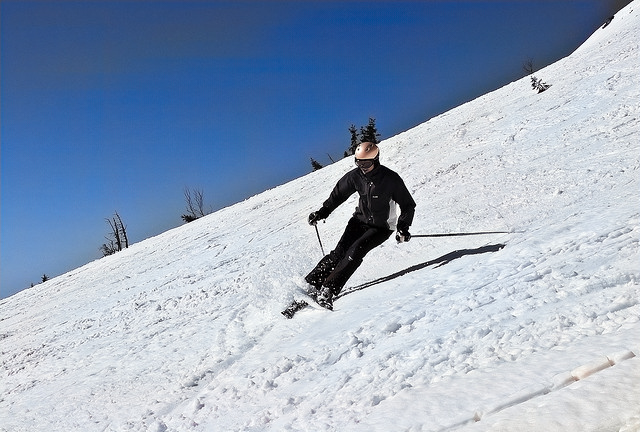}
    & 
    \includegraphics[width=0.2\textwidth]{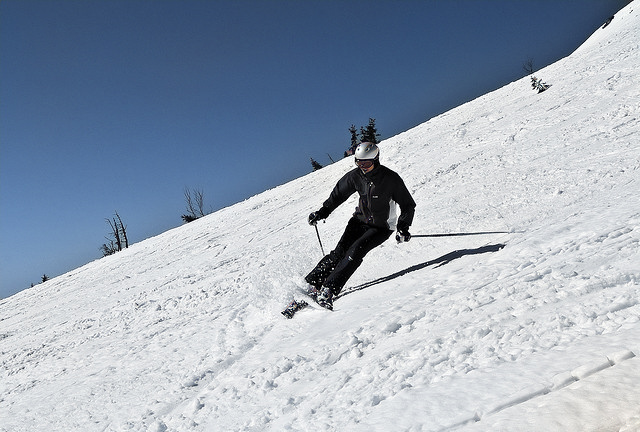}
\\
    \includegraphics[width=0.2\textwidth]{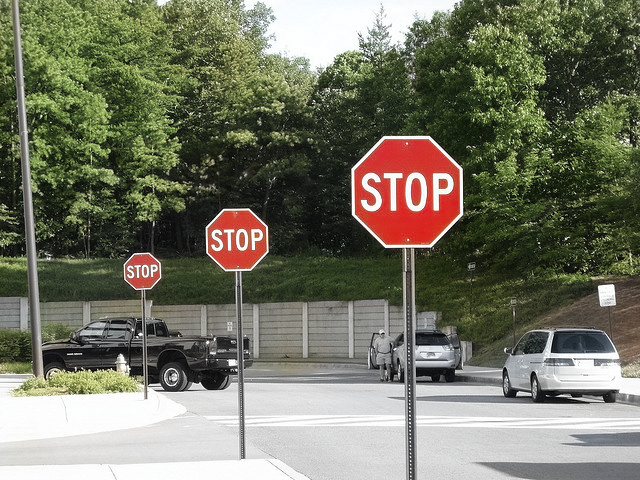}
& 
    \includegraphics[width=0.2\textwidth]{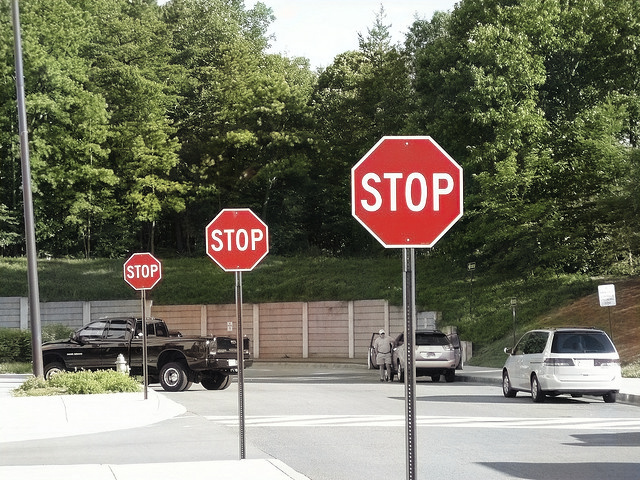}
    & 
    \includegraphics[width=0.2\textwidth]{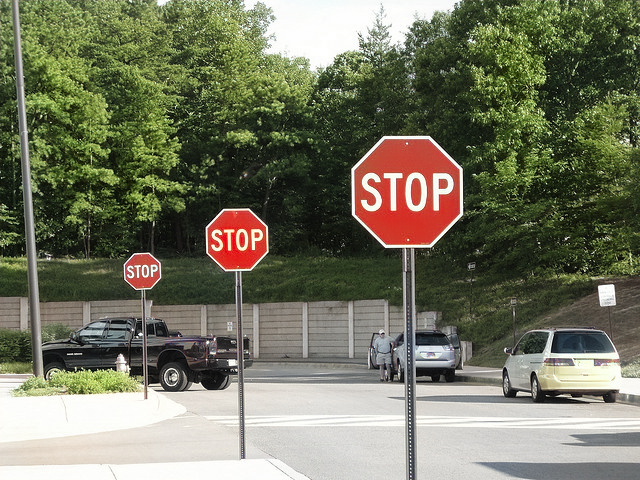}
& 
    \includegraphics[width=0.2\textwidth]{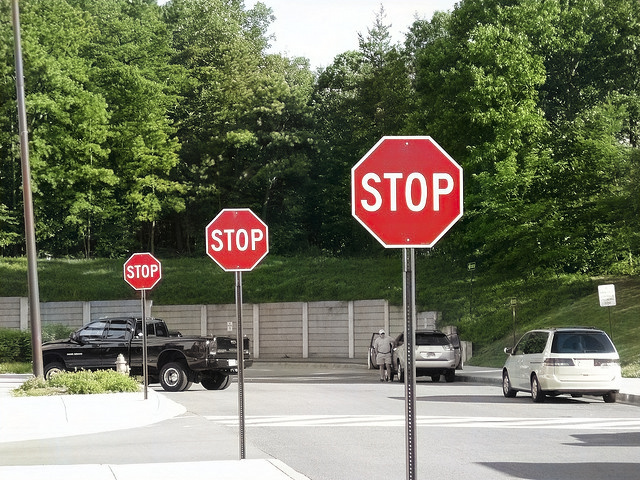}
    & 
    \includegraphics[width=0.2\textwidth]{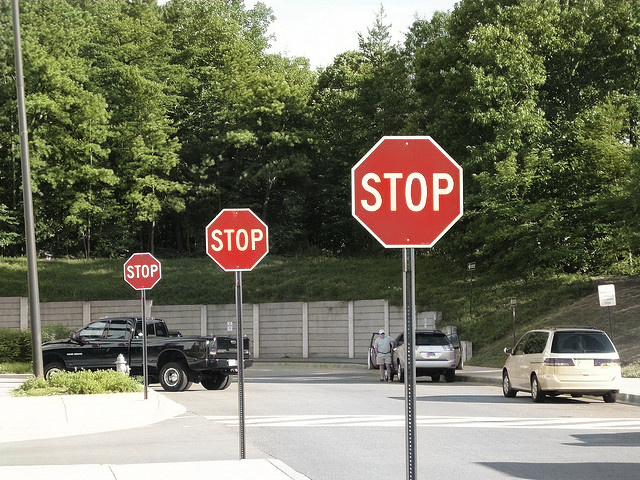}
\\
    \includegraphics[width=0.2\textwidth]{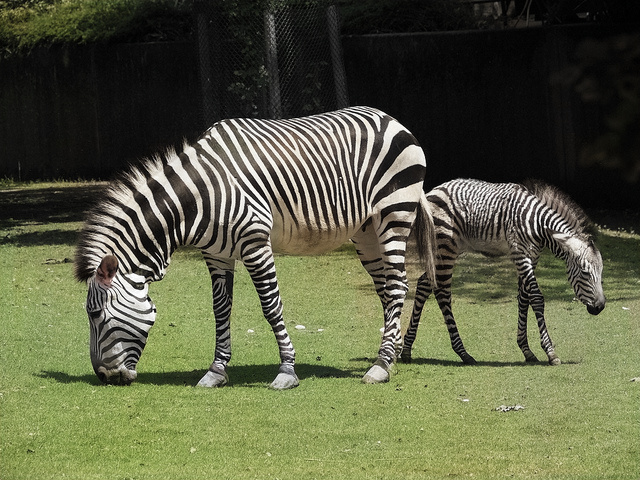}
& 
    \includegraphics[width=0.2\textwidth]{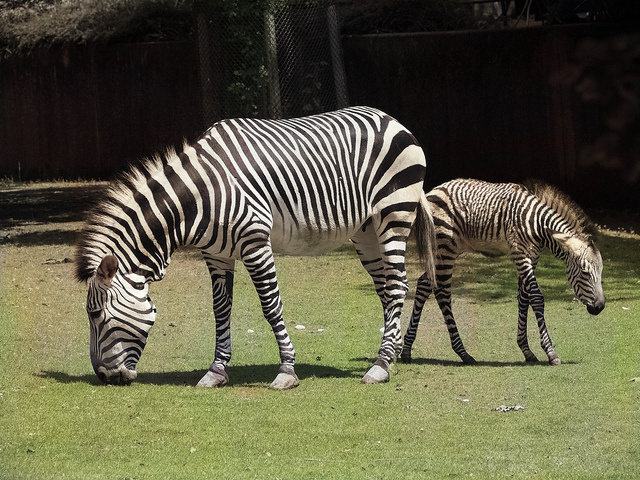}
    & 
    \includegraphics[width=0.2\textwidth]{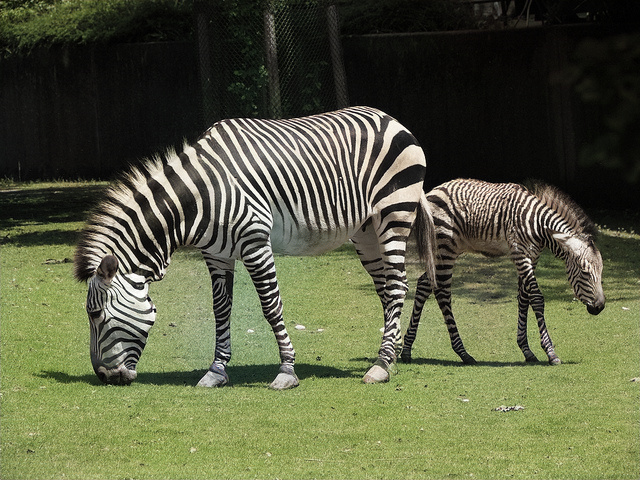}
& 
    \includegraphics[width=0.2\textwidth]{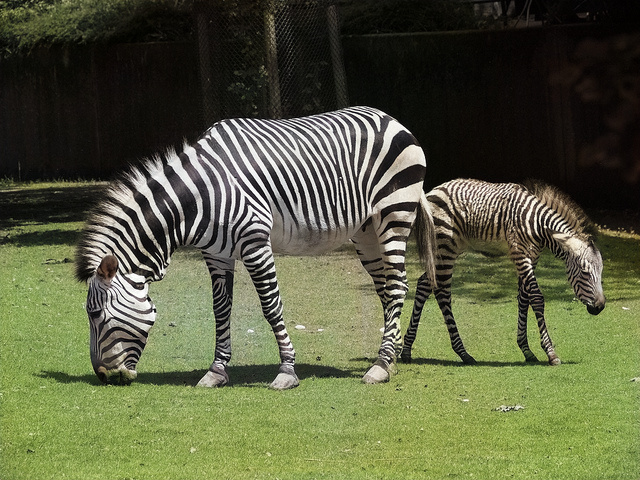}
        & 
    \includegraphics[width=0.2\textwidth]{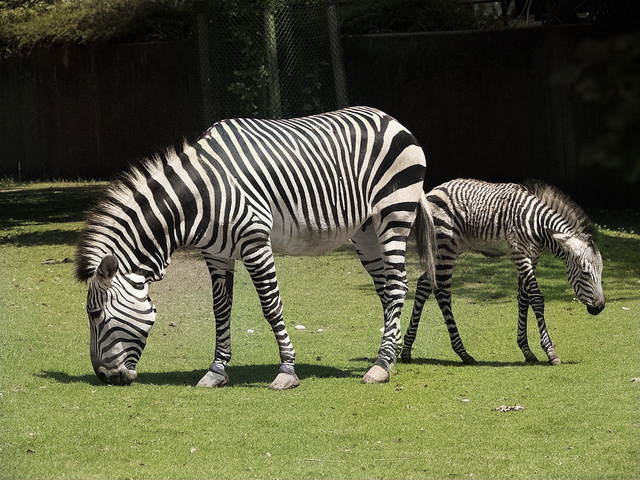}
\\
Lab-L1  & 
Lab-L2  & 
Lab-LPIPS  & 
Lab-WGAN+L2 &
Lab-WGAN+LPIPS
\\
\end{tabular}
\caption{Colorization results on images that contain objects have strong structures and that have been seen many times in the training set. Five losses are compared, namely, L1, L2, LPIPS, WGAN+L2 and WGAN+LPIPS. The used color space is Lab for all the cases.}
\label{strongstucturesLab}
\end{center}
\end{figure}

Figures~\ref{manyobjectsLab},~\ref{shinyLab} and~\ref{strongstucturesLab} show a qualitative experimental comparison of the five losses, namely, L1, L2, VGG-based LPIPS, WGAN+L2 and WGAN+VGG-based LPIPS.
In all cases, the models were trained on the Lab color space.
Still, we recall that any model based on VGG-based LPIPS loss requires to convert the predicted image to the RGB color space in a differentiable way (\emph{i.e.},~with Kornia~\cite{riba2020kornia}).

In Figure~\ref{manyobjectsLab}, we can see some results obtained for each of the studied losses in images with multiple objects. We can observe that each loss brings slightly different colors to objects.
Overall, VGG-based LPIPS and WGAN losses generate shinier and more colorful images (it can be seen, for instance in the sky, grass, and vegetables), although we can observe colorful examples in the case of the L2 loss in the example of the flowers or vegetables.
However, WGAN hallucinates more unrealistic colors as can be seen on the table or the wall on the image with a flower of the last row of Figure~\ref{manyobjectsLab}.
This effect can be reduced by improving architecture and semantic features (\emph{e.g.},~\cite{vitoria2020chromagan}) or by introducing spatial localization (\emph{e.g.},~\cite{su2020instance}).
Besides, by comparing the two last columns obtained with the models trained with the adversarial strategy WGAN combined with, respectively, the L2 or the VGG-based LPIPS, one can observe that WGAN+VGG-based LPIPS tends to homogenize colors (\emph{e.g.},~some of the balloons take similar color to the sky on the second row; the flowers on the fifth have grayish colors, more similar to the wall).
WGAN+VGG-based LPIPS also tends to have less bleeding than WGAN+L2.

The generation of more vivid colors with VGG-based LPIPS and WGAN losses in also visible on Figure~\ref{shinyLab}.
The grass and bushes are more green and look more natural.
However, none of the losses give consistency to all the limbs of the tennis player on the first row (\emph{e.g.},~the right leg).

Figure~\ref{strongstucturesLab} shows results on objects, here zebra and stop sign, with strong contours that were highly present in the training set.
The colorization of this object is impressive for any loss.
None of the losses manage to properly colorize the person near the center car on the first row.
This type of examples, could be improved by learning high-level semantics on the image content.

\begin{figure}[t] \begin{center}
\begin{tabular}{ccccc}
    \includegraphics[width=0.2\textwidth]{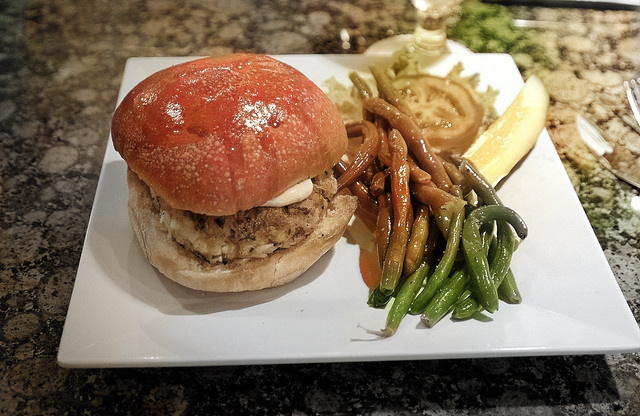}
& 
    \includegraphics[width=0.2\textwidth]{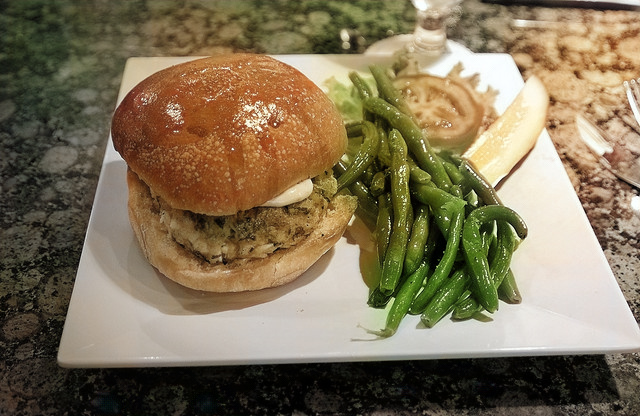}
    & 
    \includegraphics[width=0.2\textwidth]{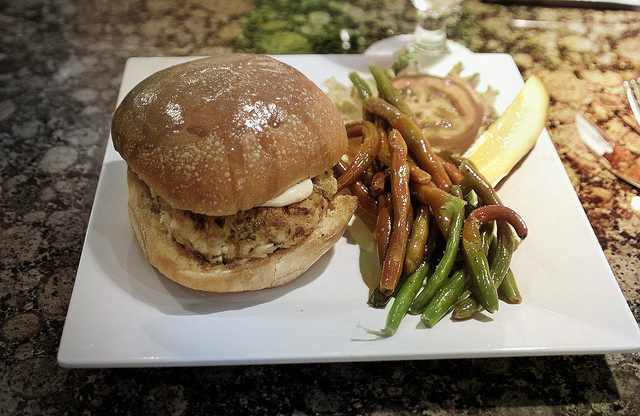}
& 
    \includegraphics[width=0.2\textwidth]{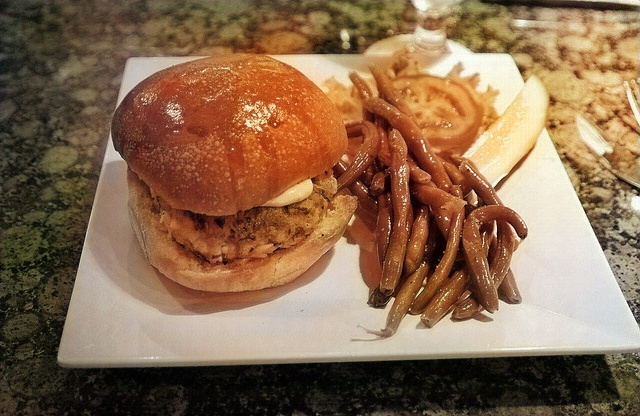}
& 
    \includegraphics[width=0.2\textwidth]{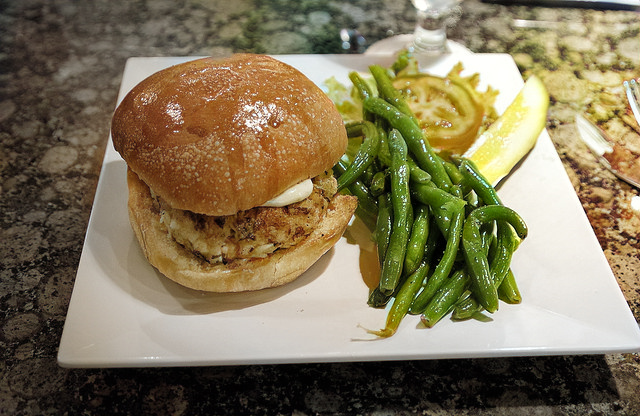}
\\
    \includegraphics[width=0.2\textwidth]{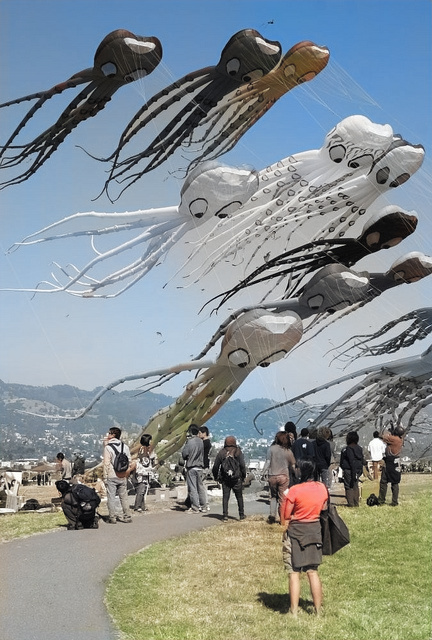}
& 
    \includegraphics[width=0.2\textwidth]{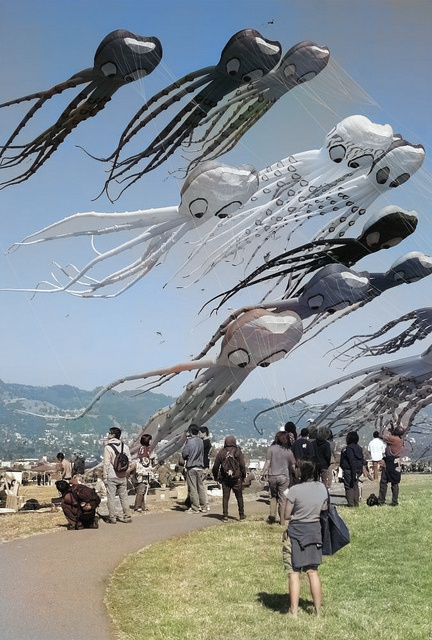}
    & 
    \includegraphics[width=0.2\textwidth]{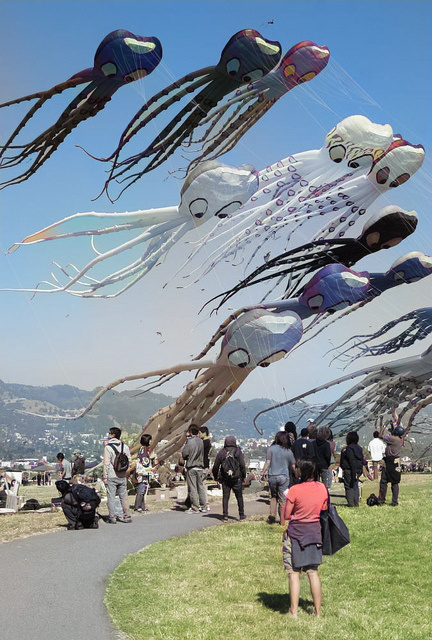}
& 
    \includegraphics[width=0.2\textwidth]{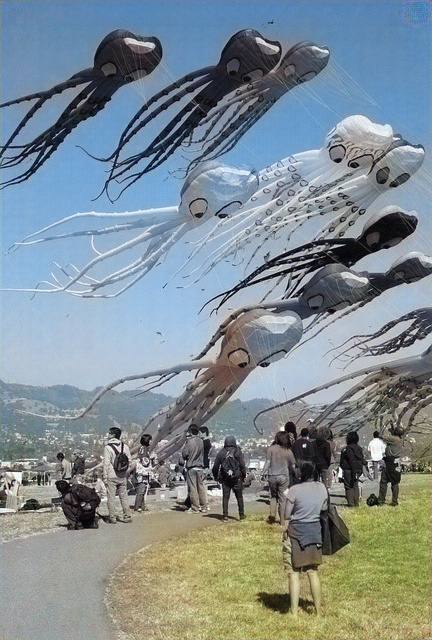}
&
\includegraphics[width=0.2\textwidth]{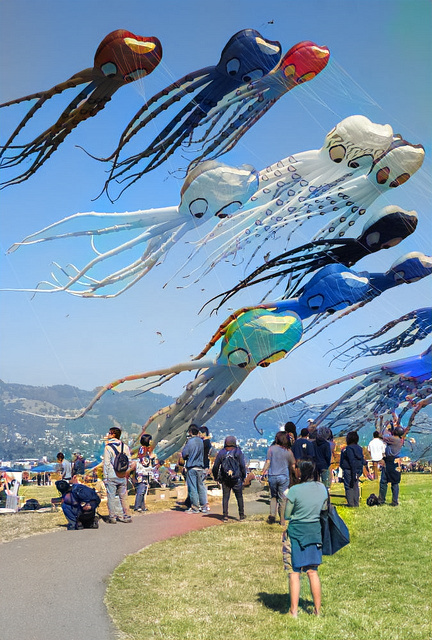}
\\
    \includegraphics[width=0.2\textwidth]{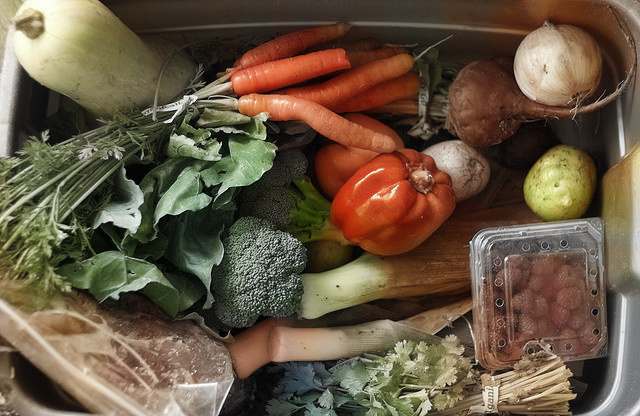}
& 
    \includegraphics[width=0.2\textwidth]{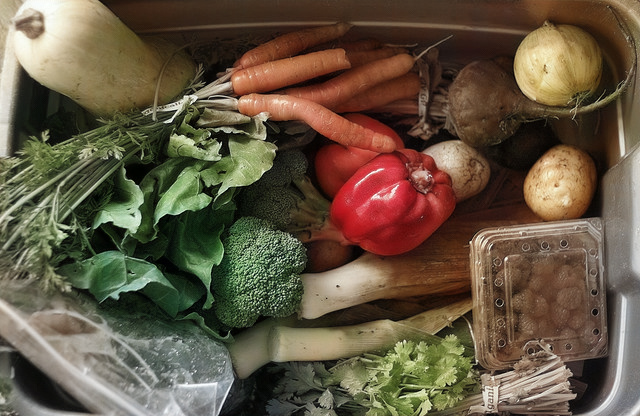}
    & 
    \includegraphics[width=0.2\textwidth]{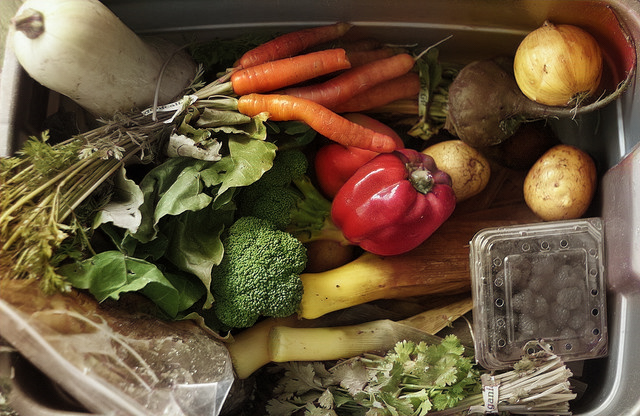}
& 
    \includegraphics[width=0.2\textwidth]{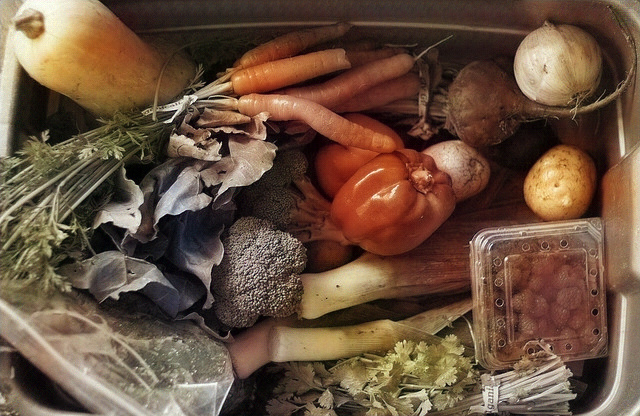}
& 
    \includegraphics[width=0.2\textwidth]{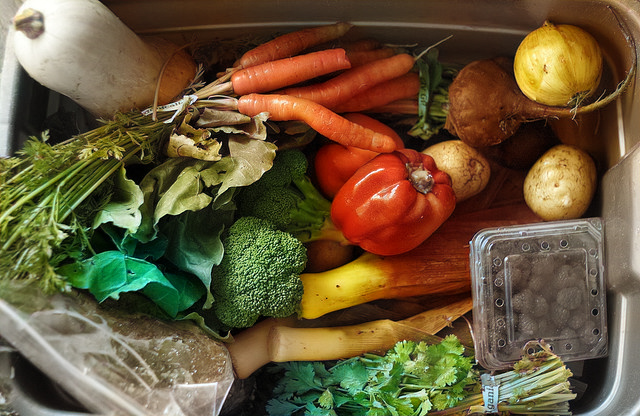}
\\
    \includegraphics[width=0.2\textwidth]{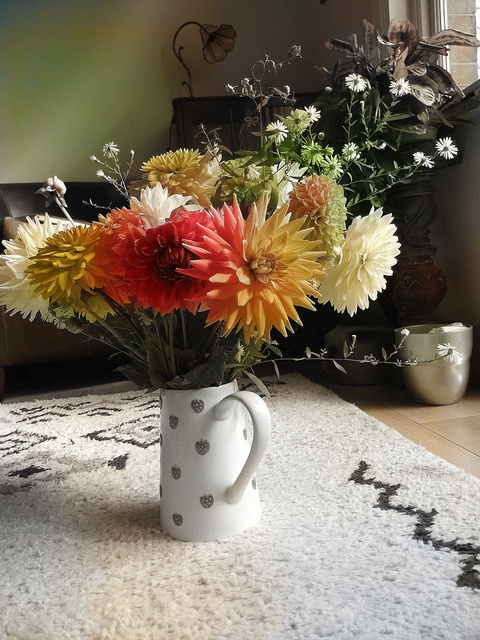}
& 
    \includegraphics[width=0.2\textwidth]{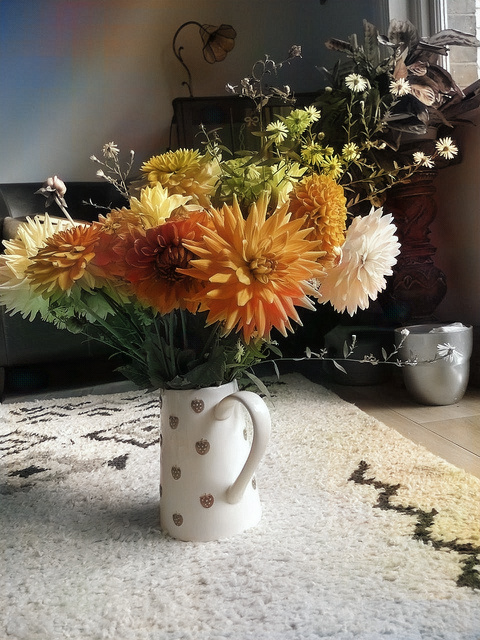}
    & 
    \includegraphics[width=0.2\textwidth]{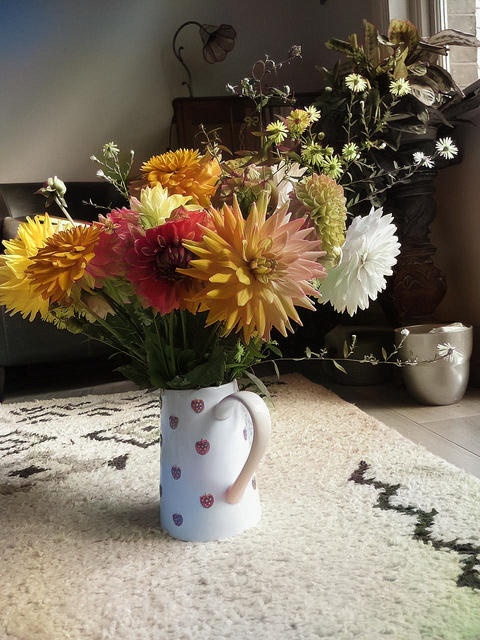}
& 
    \includegraphics[width=0.2\textwidth]{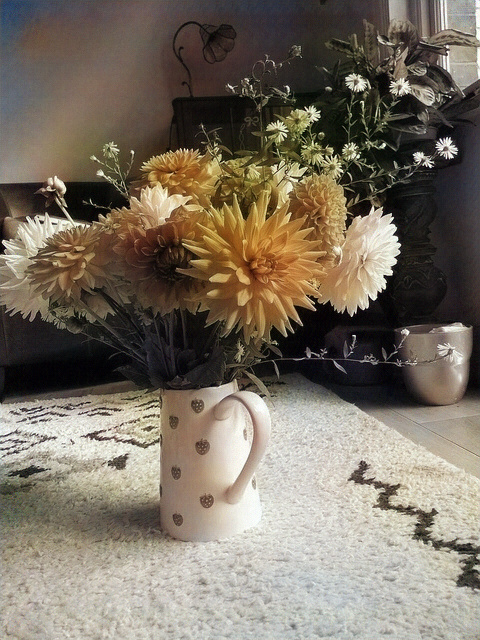}
& 
    \includegraphics[width=0.2\textwidth]{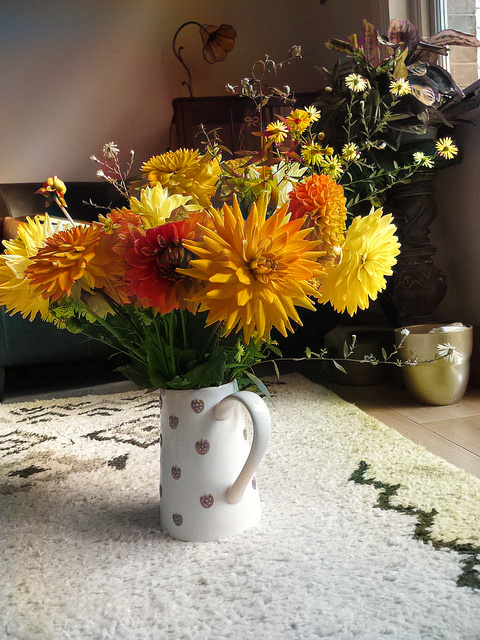}
\\
 RGB-L1  & 
 RGB-L2  & 
 Lab-LPIPS  & 
Lab-WGAN+L2 & 
Lab-WGAN+LPIPS
\\
 \end{tabular}
    \caption{Examples where multiple objects are in the same image. Four losses are compared, namely, L1, L2, LPIPS, WGAN+L2, and WGAN+LPIPS perceptual. 
    The used color space is RGB for all the cases (that is, the model estimates three RGB color channels).}\label{manyobjectsRGB}
 \end{center} \end{figure}

\begin{figure}[t] \begin{center}
\begin{tabular}{ccccc}
    \includegraphics[width=0.2\textwidth]{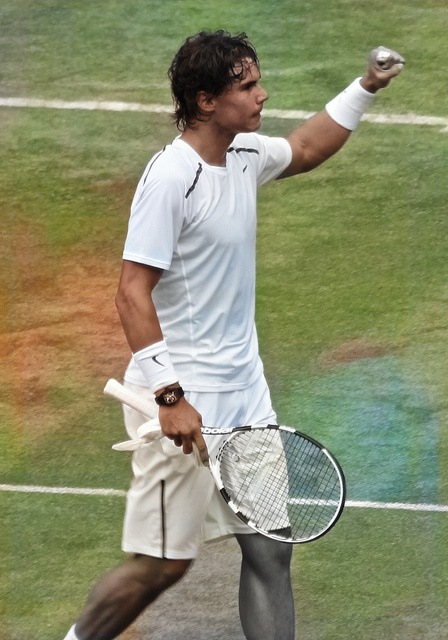}
& 
    \includegraphics[width=0.2\textwidth]{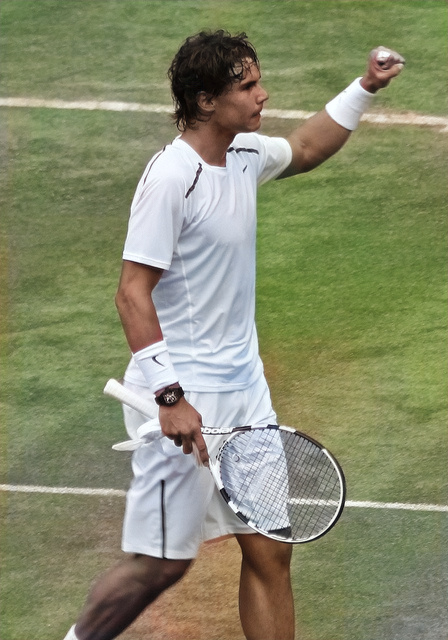}
    & 
    \includegraphics[width=0.2\textwidth]{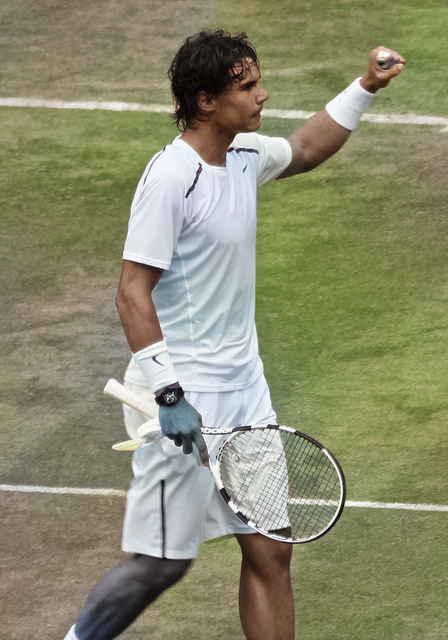}
& 
    \includegraphics[width=0.2\textwidth]{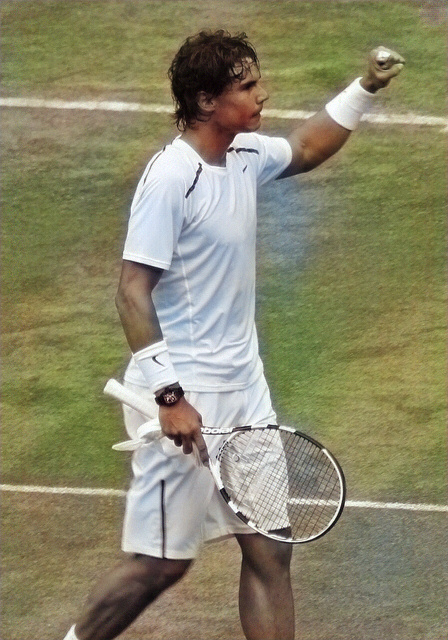}
& 
    \includegraphics[width=0.2\textwidth]{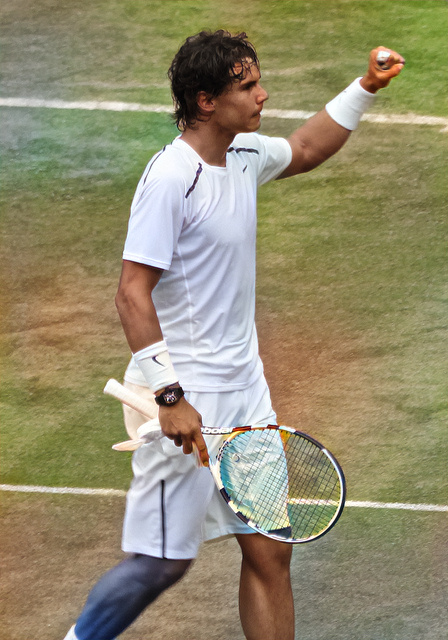}
\\
    \includegraphics[width=0.2\textwidth]{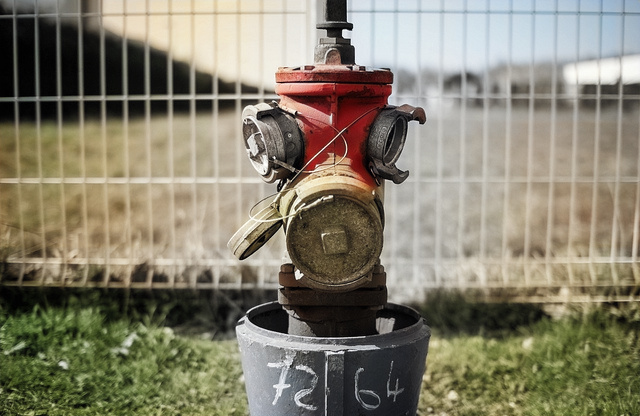}
& 
    \includegraphics[width=0.2\textwidth]{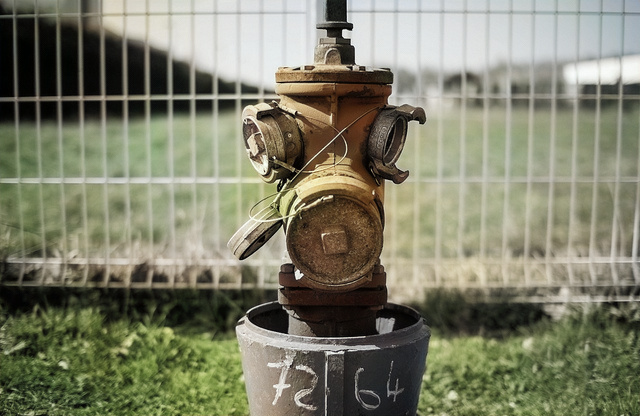}
    & 
    \includegraphics[width=0.2\textwidth]{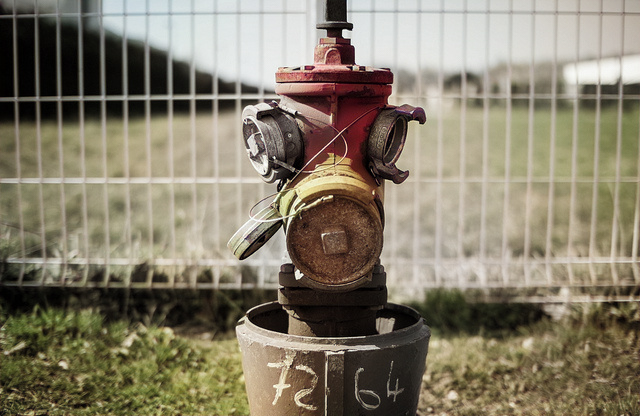}
& 
    \includegraphics[width=0.2\textwidth]{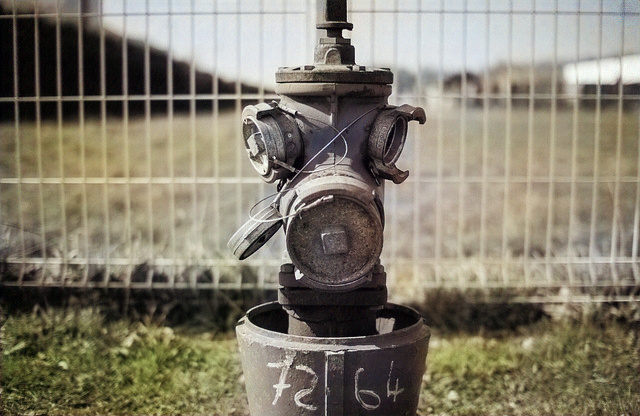}
& 
    \includegraphics[width=0.2\textwidth]{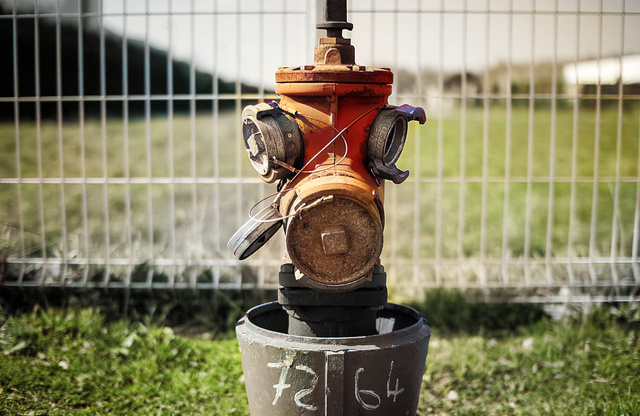}
\\
    \includegraphics[width=0.2\textwidth]{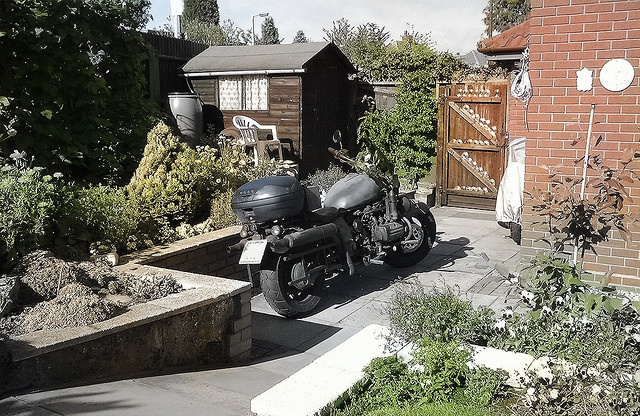}
& 
    \includegraphics[width=0.2\textwidth]{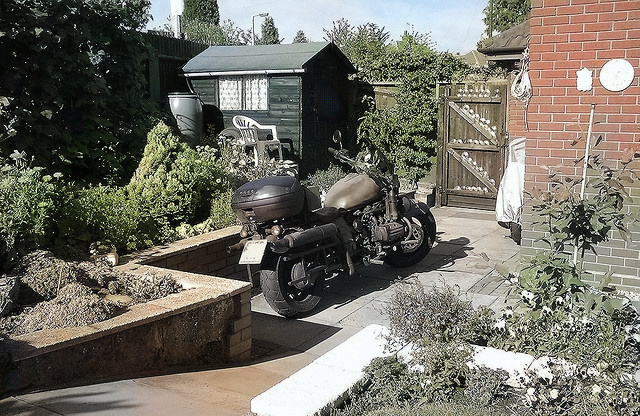}
    & 
    \includegraphics[width=0.2\textwidth]{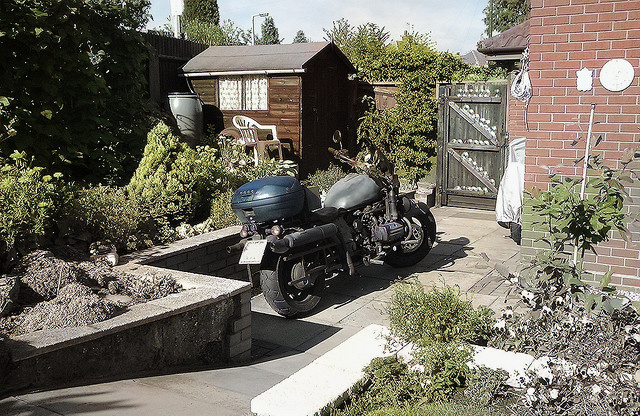}
& 
    \includegraphics[width=0.2\textwidth]{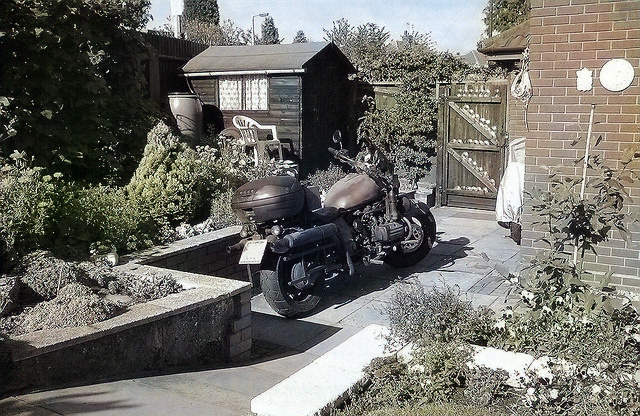}
& 
    \includegraphics[width=0.2\textwidth]{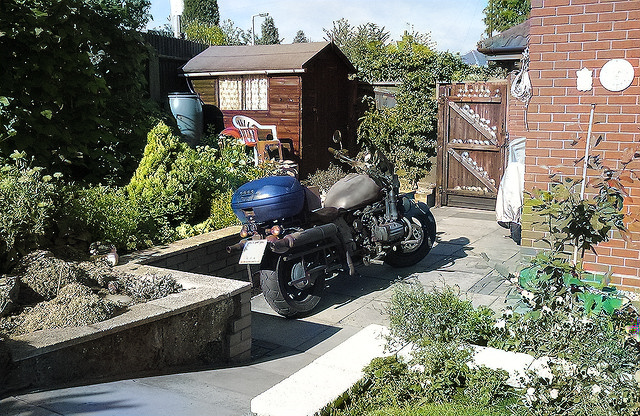}
\\
 RGB-L1  & 
 RGB-L2  & 
 Lab-LPIPS  & 
Lab-WGAN+L2 & 
Lab-WGAN+LPIPS
\\
 \end{tabular}
    \caption{Examples to evaluate shyniness of the results. Four losses are compared, namely, L1, L2, LPIPS, WGAN+L2, and WGAN+LPIPS perceptual. 
    The used color space is RGB for all the cases.}\label{shinyRGB}
 \end{center} \end{figure}

\begin{figure}[t]
\begin{center}
\begin{tabular}{ccccc}
\includegraphics[width=0.2\textwidth]{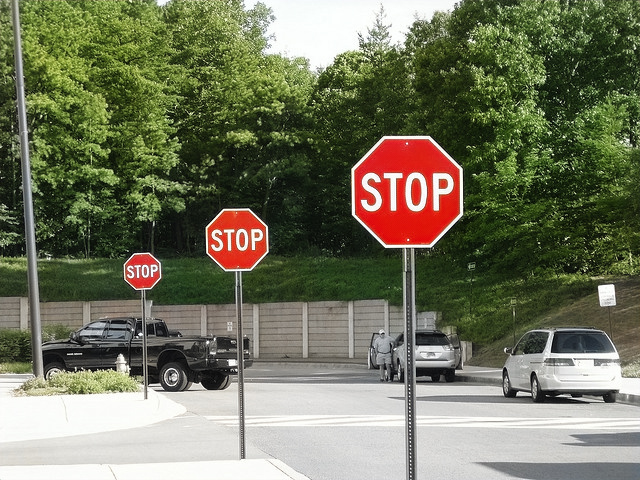} & 
\includegraphics[width=0.2\textwidth]{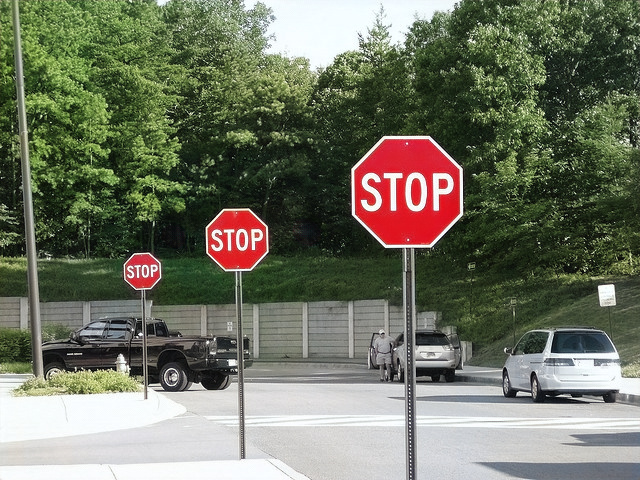} & 
\includegraphics[width=0.2\textwidth]{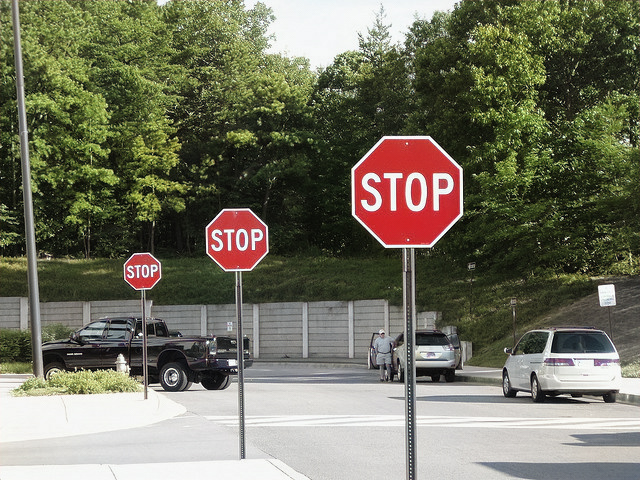} & 
\includegraphics[width=0.2\textwidth]{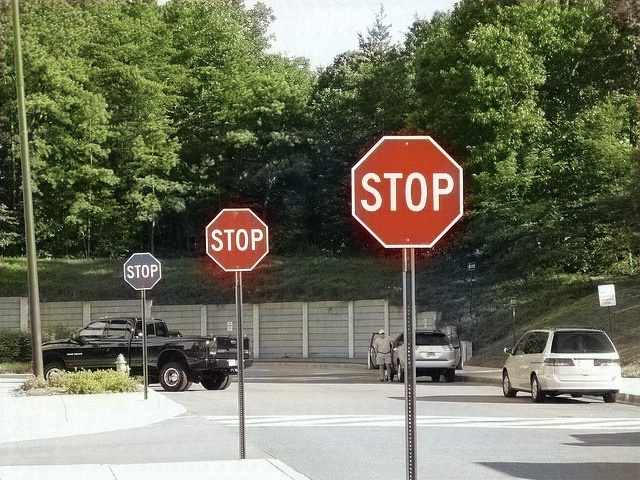} & 
\includegraphics[width=0.2\textwidth]{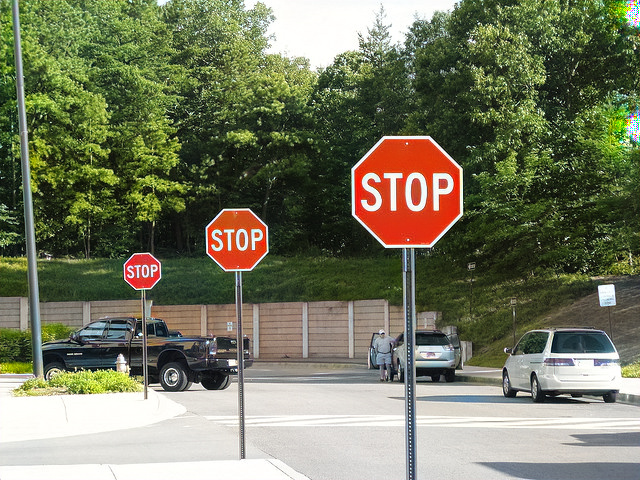} \\
\includegraphics[width=0.2\textwidth]{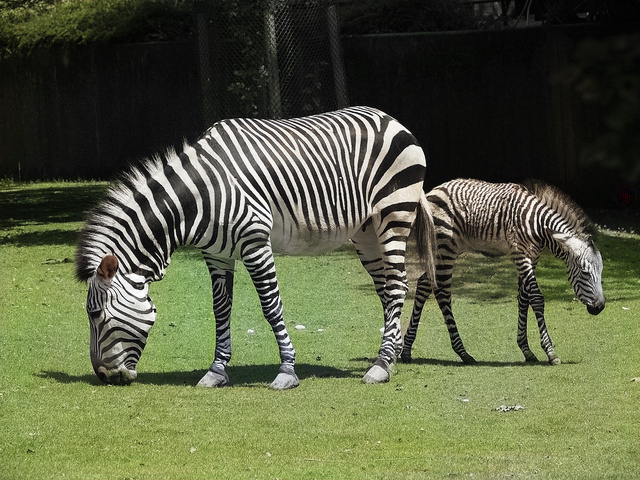} & 
\includegraphics[width=0.2\textwidth]{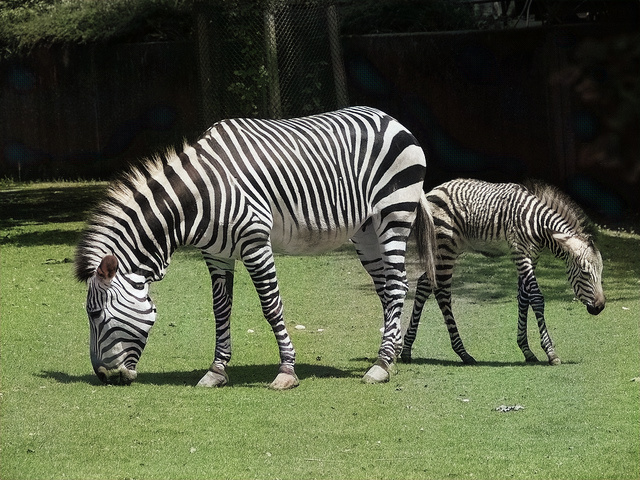} & 
\includegraphics[width=0.2\textwidth]{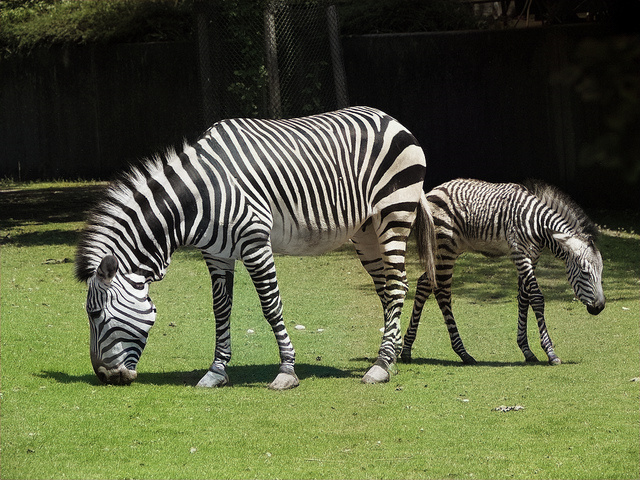} & 
\includegraphics[width=0.2\textwidth]{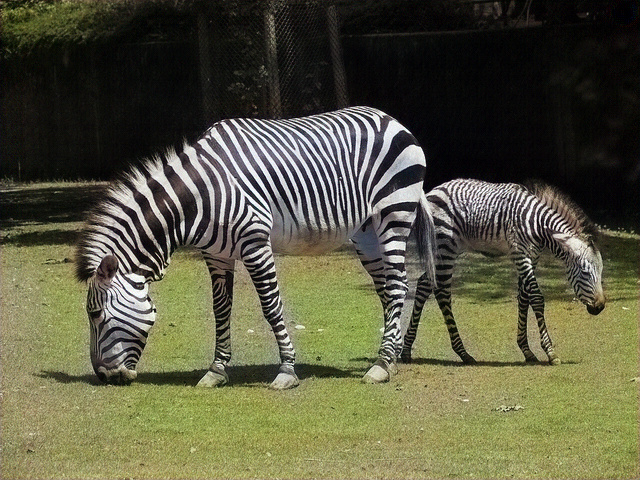} & 
\includegraphics[width=0.2\textwidth]{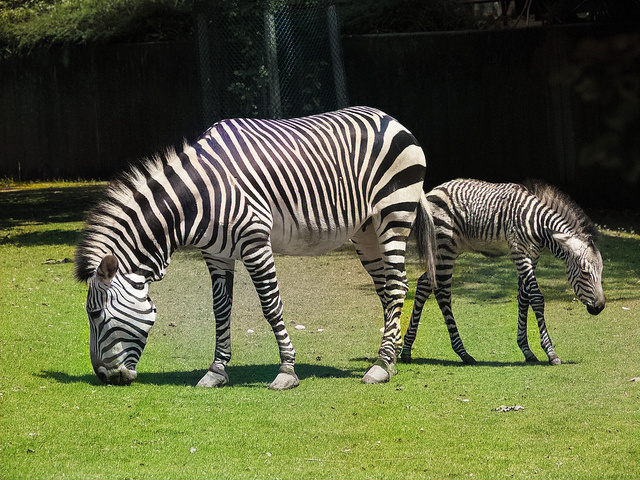} \\
RGB-L1  & RGB-L2  & RGB-LPIPS & RGB-WGAN+L2 & RGB-WGAN+LPIPS \\
\end{tabular}
\caption{Colorization results on images that contain objects have strong structures and that have been seen many times in the training set. Four losses are compared, namely, L1, L2, LPIPS, and WGAN+L2. 
The used color space is RGB for all the cases.}
\label{strongstucturesRGB}
\end{center}
\end{figure}

Figures~\ref{manyobjectsRGB},~\ref{shinyRGB} and~\ref{strongstucturesRGB} show an additional experimental comparison of five losses, namely, L1, L2, VGG-based LPIPS, WGAN+L2, and WGAN+VGG-based LPIPS, but when the network is trained to learn the three RGB color channels for all the cases.
For these test images, more realistic and consistent results are obtained in general for this configuration.
Let us notice from the results in these three figures that more colorful images are obtained compared to the ones of Figures~\ref{manyobjectsLab},~\ref{shinyLab} and~\ref{strongstucturesLab}, although less textured.
Further analysis on the influence of the chosen color space can be found in the other chapter \emph{Influence of Color Spaces for Deep Learning Image Colorization} \cite{ballester2022influence}.

\section{Generalization to Archive Images}
\label{sec:archive}

\begin{figure}[t]
\begin{center}
\begin{tabular}{ccccc}
    \includegraphics[width=0.2\textwidth]{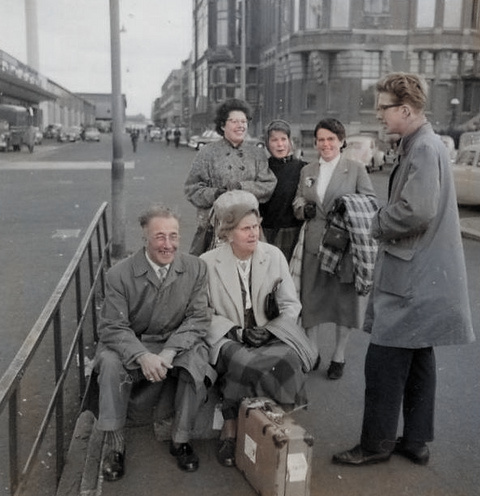}
& 
    \includegraphics[width=0.2\textwidth]{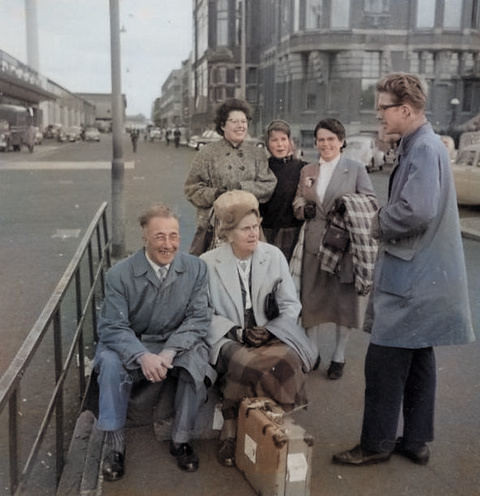}
& 
    \includegraphics[width=0.2\textwidth]{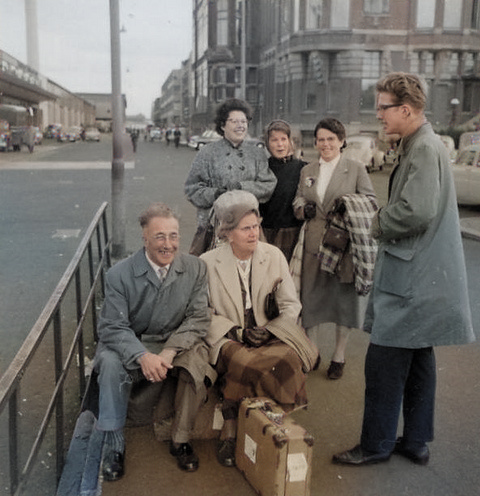}
& 
    \includegraphics[width=0.2\textwidth]{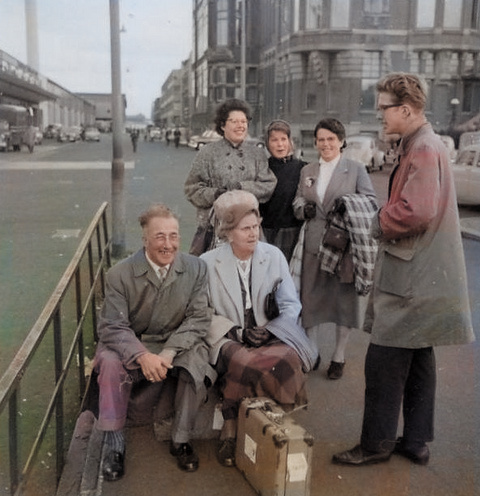}
& 
    \includegraphics[width=0.2\textwidth]{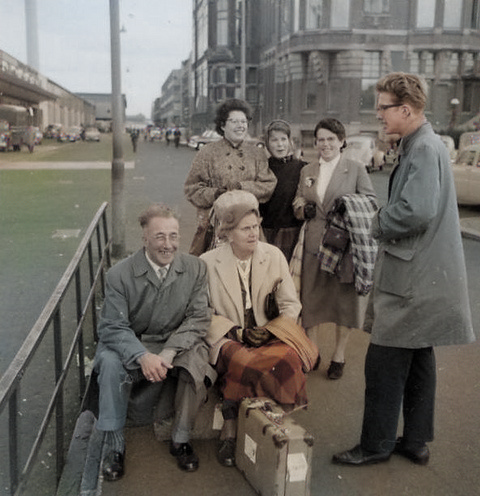}
\\
    \includegraphics[width=0.2\textwidth]{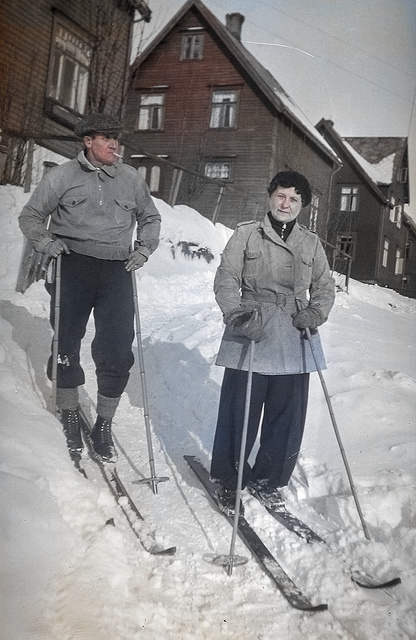}
& 
    \includegraphics[width=0.2\textwidth]{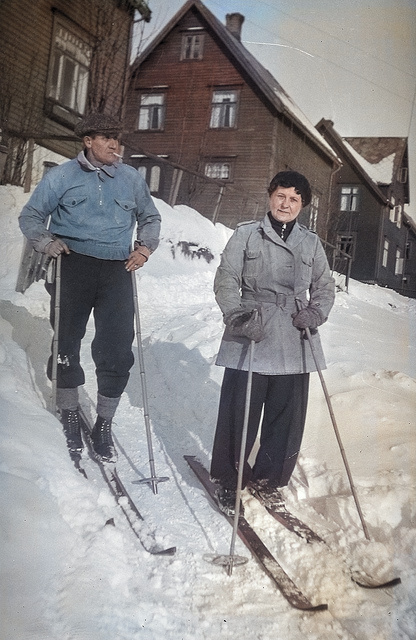}
& 
    \includegraphics[width=0.2\textwidth]{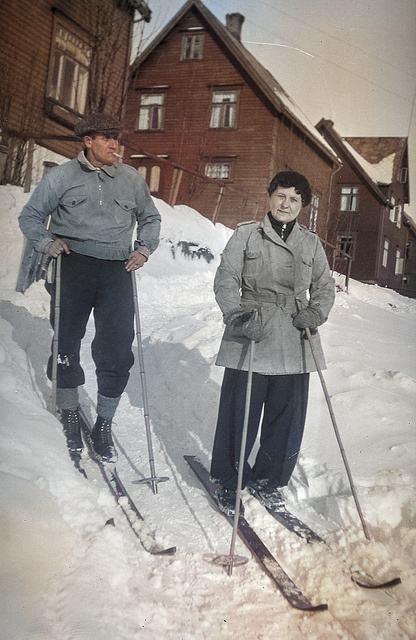}
& 
    \includegraphics[width=0.2\textwidth]{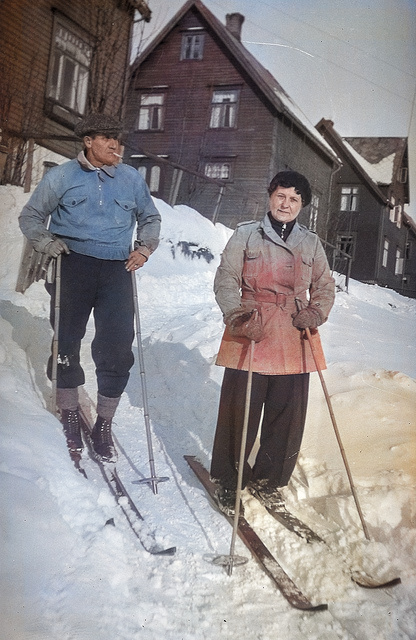}
& 
    \includegraphics[width=0.2\textwidth]{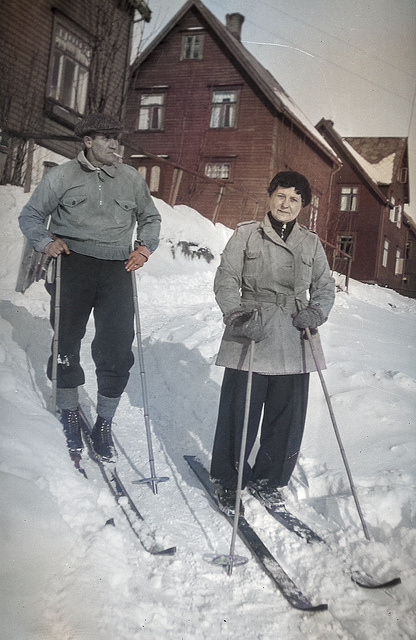}
\\
    \includegraphics[width=0.2\textwidth]{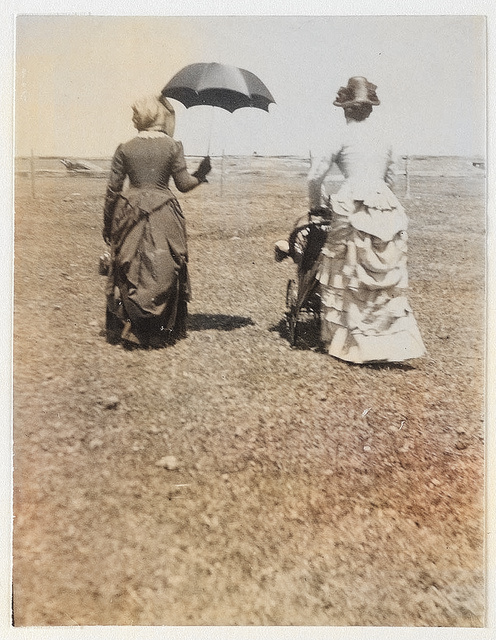}
& 
    \includegraphics[width=0.2\textwidth]{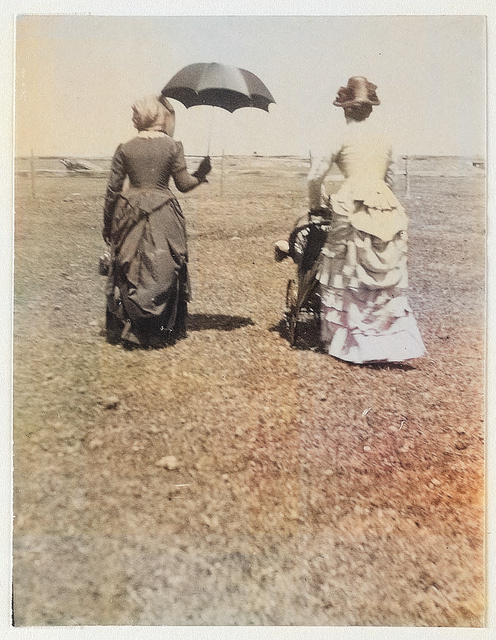}
& 
    \includegraphics[width=0.2\textwidth]{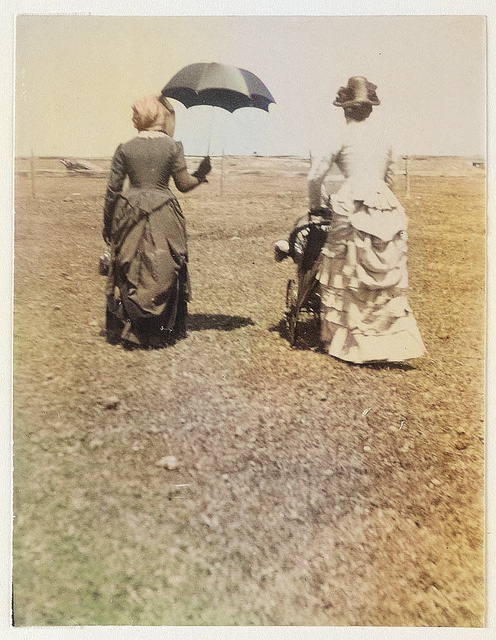}
& 
    \includegraphics[width=0.2\textwidth]{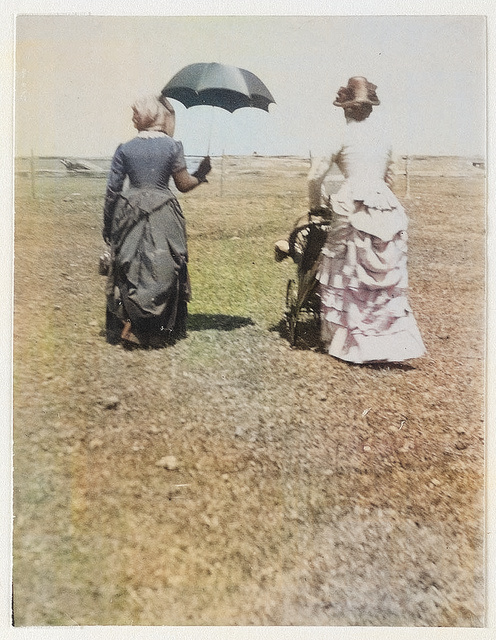}
& 
    \includegraphics[width=0.2\textwidth]{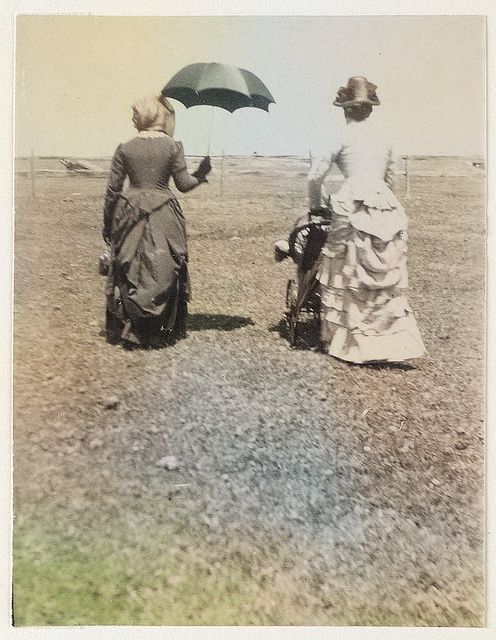}
\\
    \includegraphics[width=0.2\textwidth]{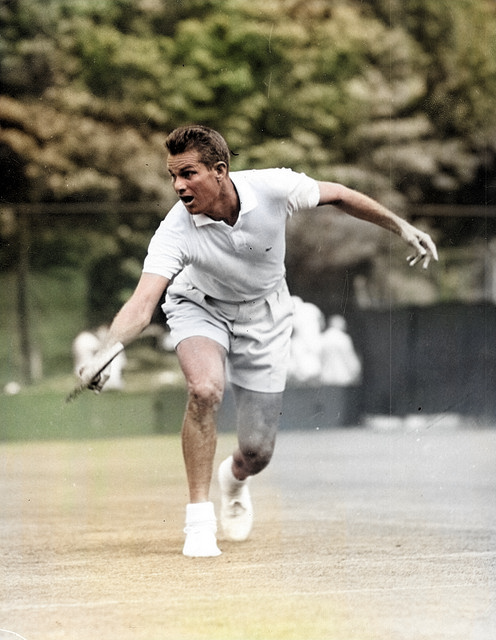}
& 
    \includegraphics[width=0.2\textwidth]{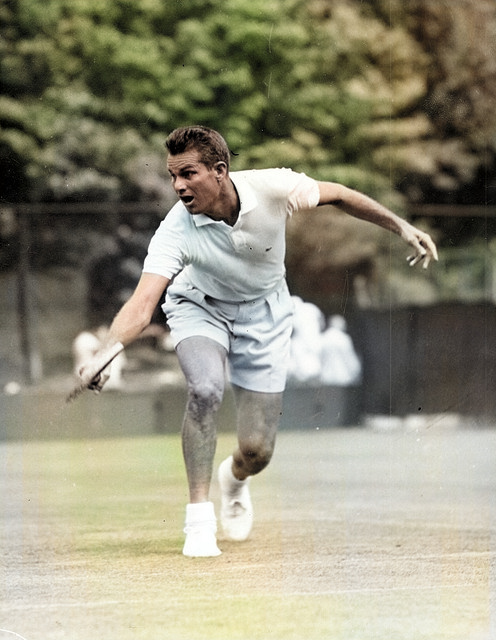}
& 
    \includegraphics[width=0.2\textwidth]{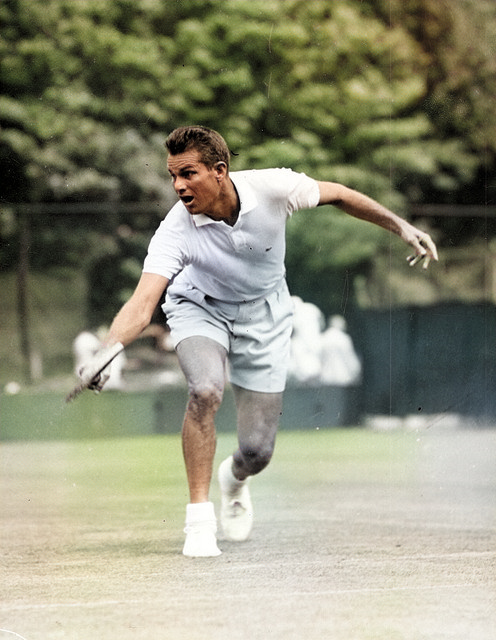}
& 
    \includegraphics[width=0.2\textwidth]{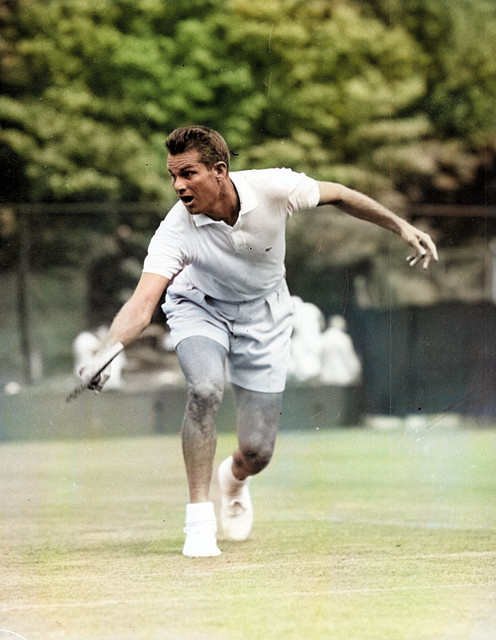}
& 
    \includegraphics[width=0.2\textwidth]{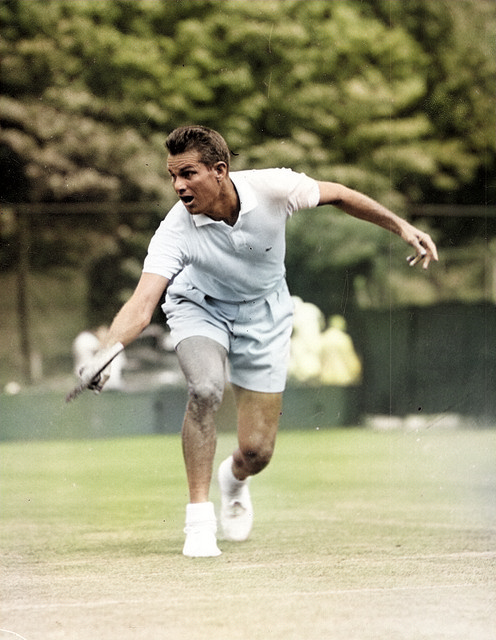}
\\
    \includegraphics[width=0.2\textwidth]{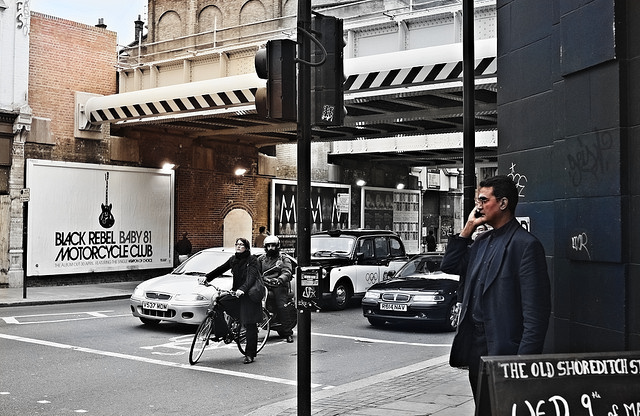}
& 
    \includegraphics[width=0.2\textwidth]{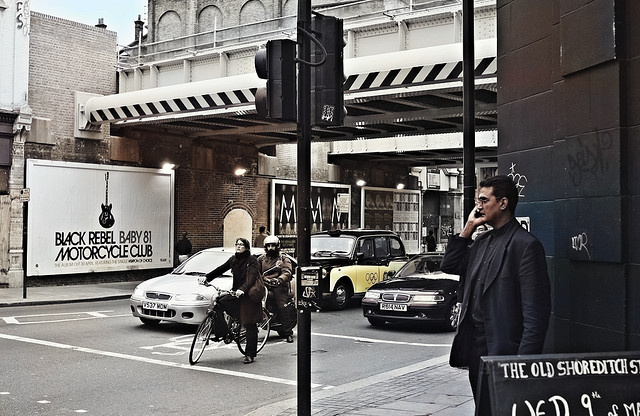}
& 
    \includegraphics[width=0.2\textwidth]{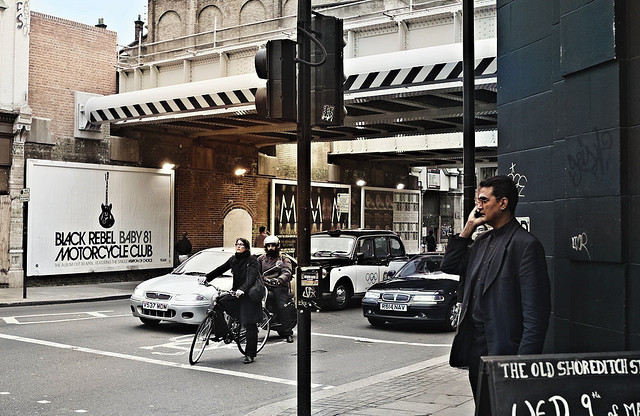}
& 
    \includegraphics[width=0.2\textwidth]{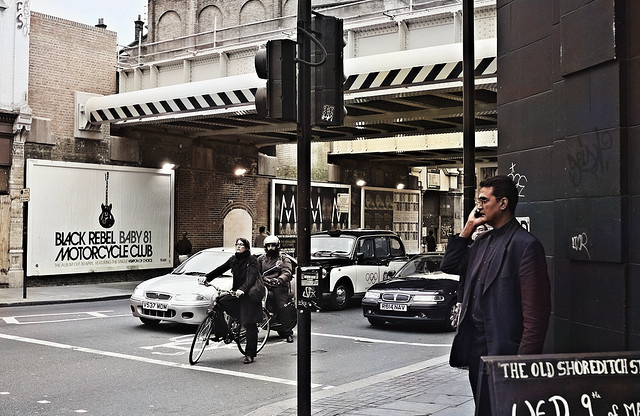}
& 
    \includegraphics[width=0.2\textwidth]{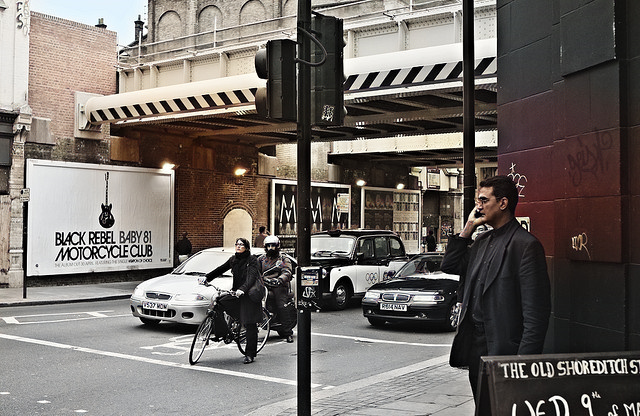}
\\
    \includegraphics[width=0.2\textwidth]{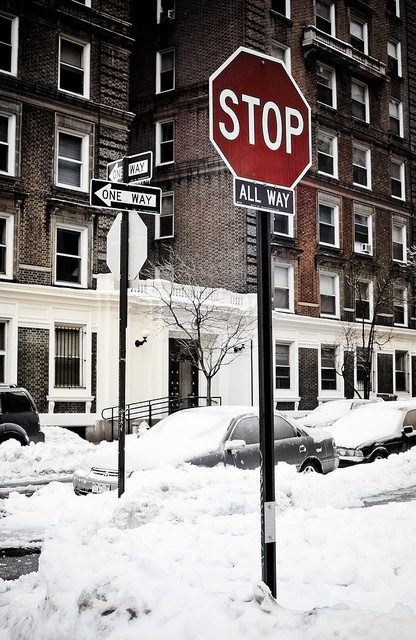}
& 
    \includegraphics[width=0.2\textwidth]{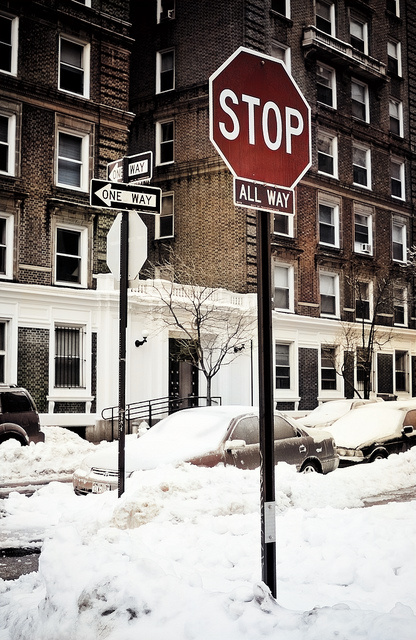}
& 
    \includegraphics[width=0.2\textwidth]{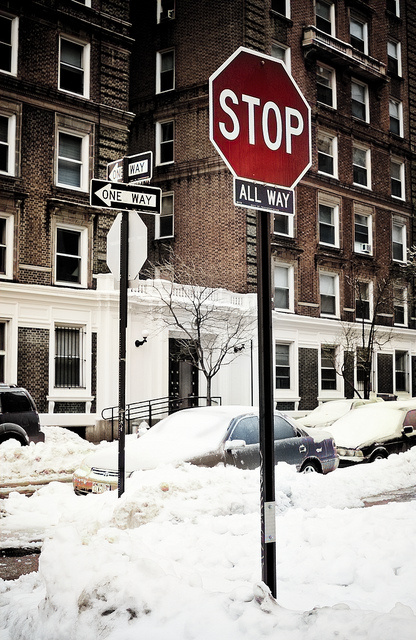}
& 
    \includegraphics[width=0.2\textwidth]{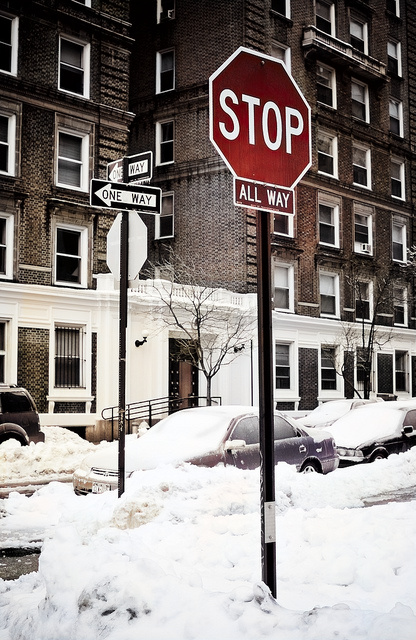}
& 
    \includegraphics[width=0.2\textwidth]{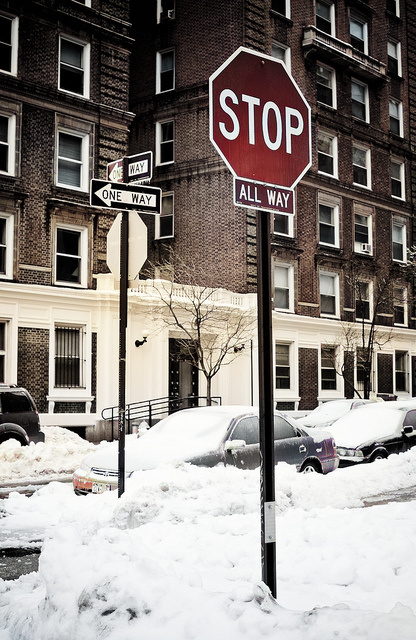}
\\
Lab-L1  & 
Lab-L2  & 
Lab-LPIPS &
Lab-WGAN+L2 &
Lab-WGAN+LPIPS
\\
\end{tabular}
\caption{Examples in Original Black and White Images. These colorization results have been obtained using the five networks trained, respectively, with  L1, L2, LPIPS, WGAN+L2 and WGAN+LPIPS losses, and learning the two ab chrominance channels.}\label{oldLab}
\end{center}
\end{figure}

\begin{figure}[t]
\begin{center}
\begin{tabular}{ccccc}
    \includegraphics[width=0.2\textwidth]{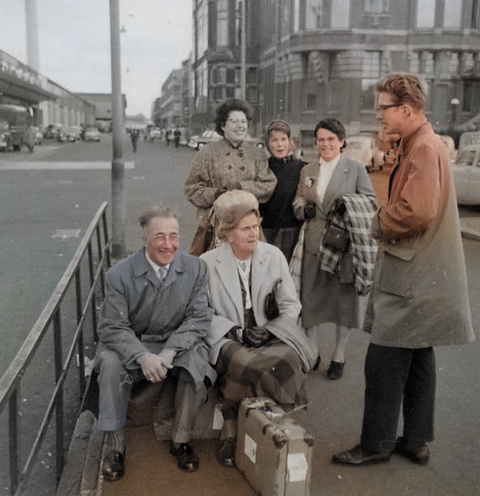}
& 
    \includegraphics[width=0.2\textwidth]{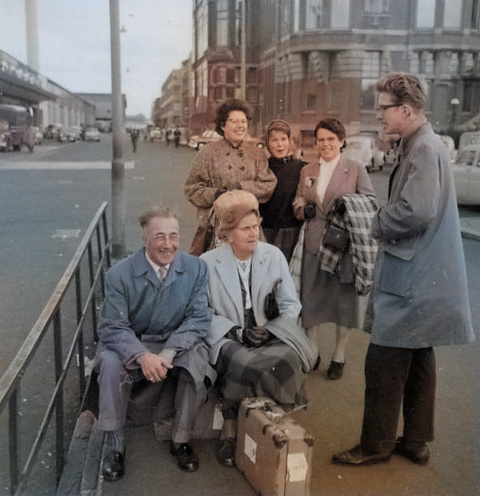}
& 
    \includegraphics[width=0.2\textwidth]{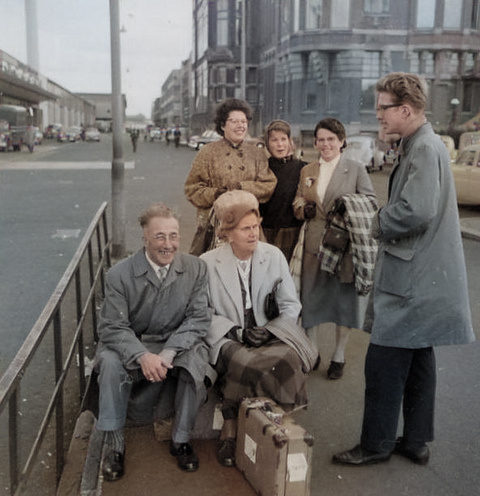}
& 
    \includegraphics[width=0.2\textwidth]{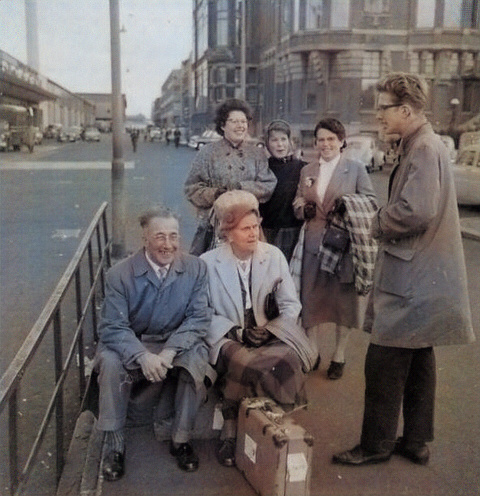}
& 
    \includegraphics[width=0.2\textwidth]{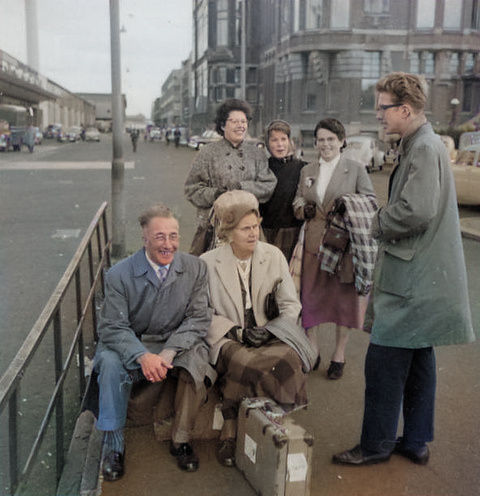}
\\
    \includegraphics[width=0.2\textwidth]{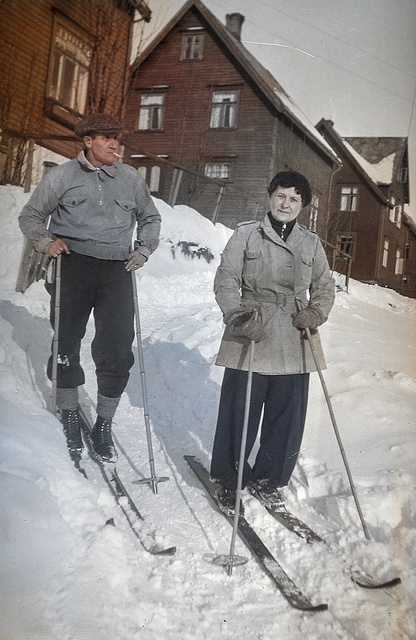}
& 
    \includegraphics[width=0.2\textwidth]{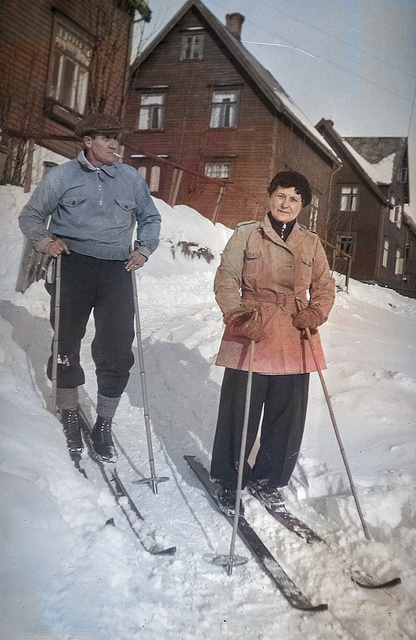}
& 
    \includegraphics[width=0.2\textwidth]{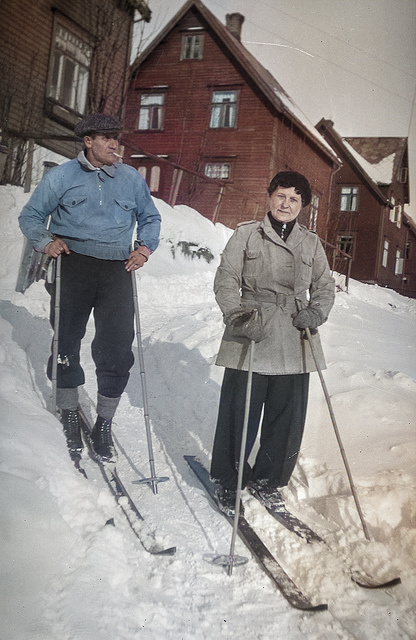}
& 
    \includegraphics[width=0.2\textwidth]{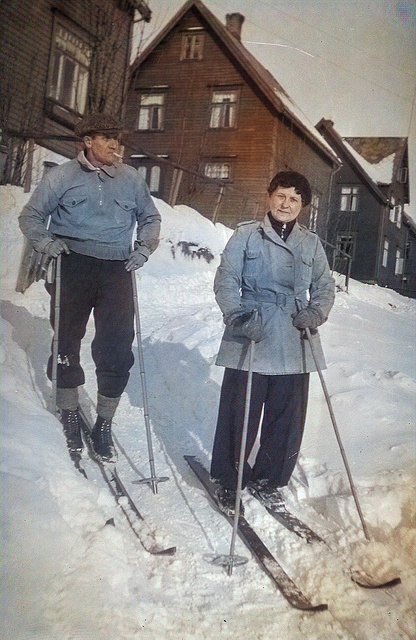}
& 
    \includegraphics[width=0.2\textwidth]{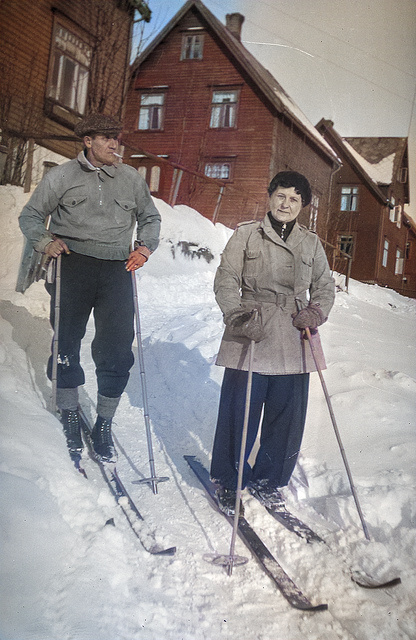}
\\
    \includegraphics[width=0.2\textwidth]{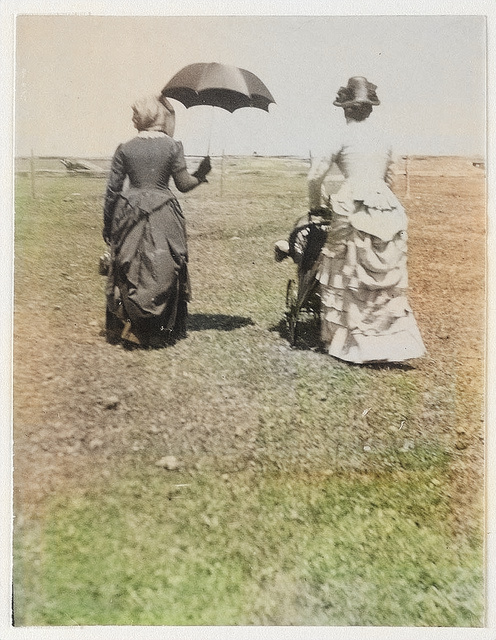}
& 
    \includegraphics[width=0.2\textwidth]{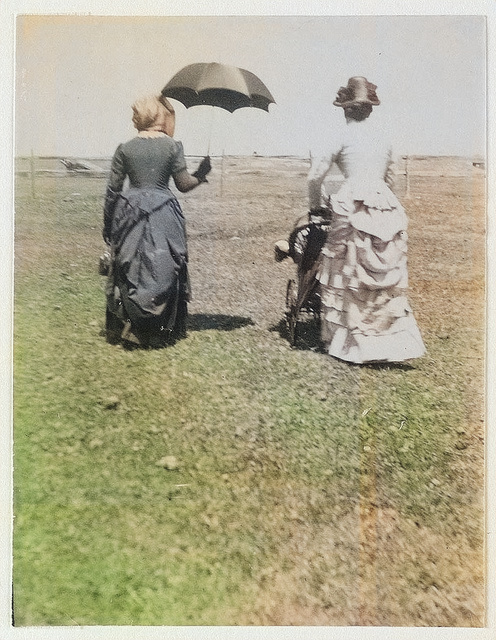}
& 
    \includegraphics[width=0.2\textwidth]{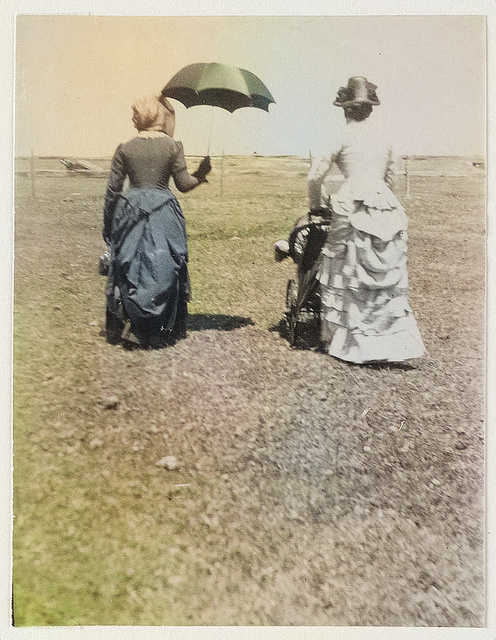}
& 
    \includegraphics[width=0.2\textwidth]{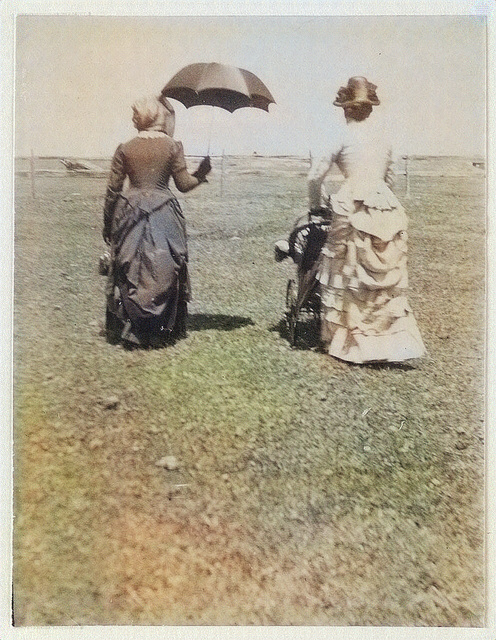}
& 
    \includegraphics[width=0.2\textwidth]{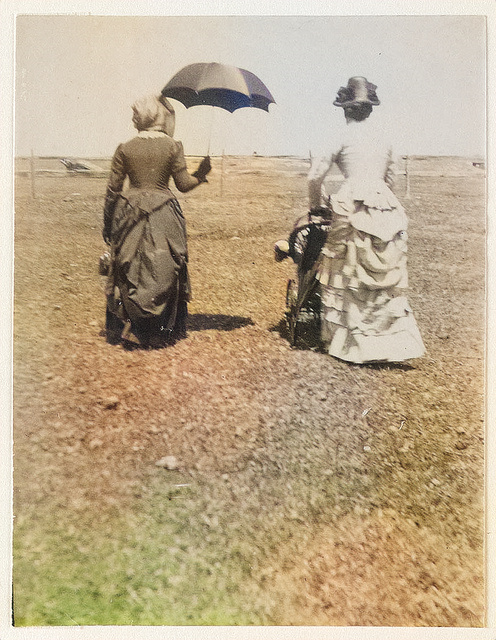}
\\
    \includegraphics[width=0.2\textwidth]{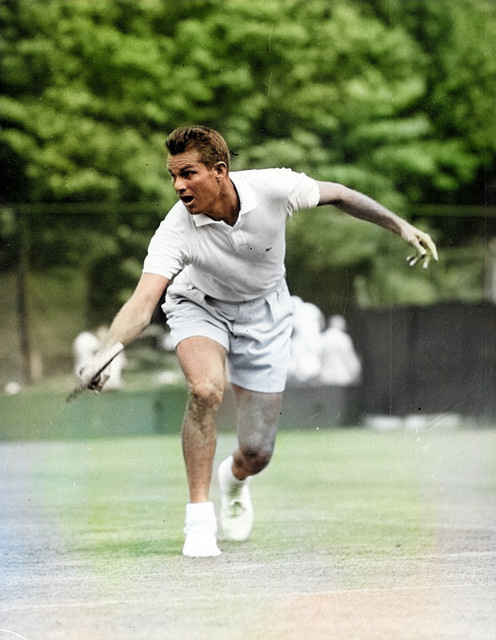}
& 
    \includegraphics[width=0.2\textwidth]{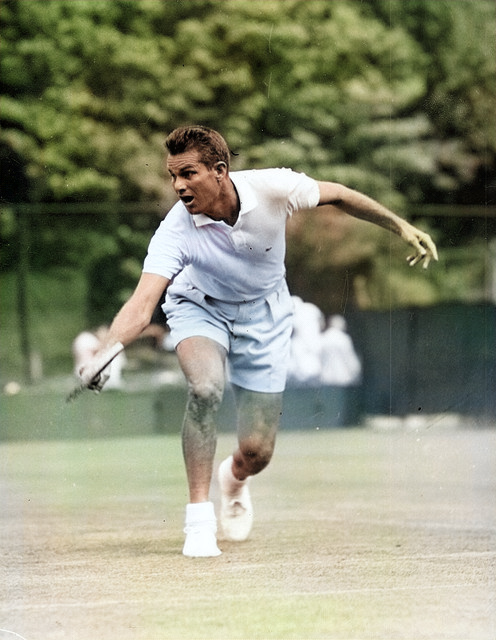}
& 
    \includegraphics[width=0.2\textwidth]{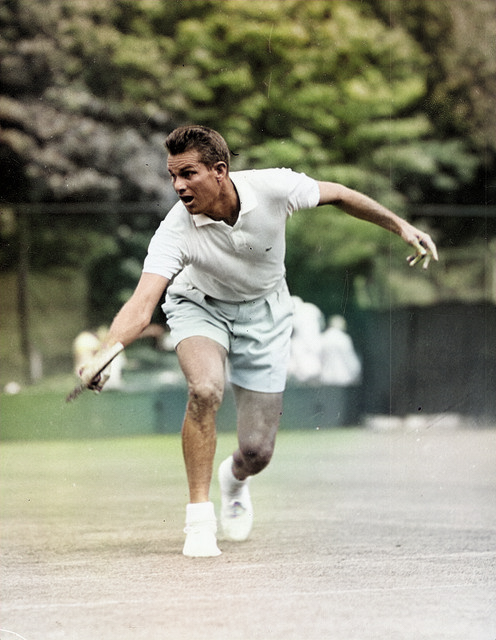}
& 
    \includegraphics[width=0.2\textwidth]{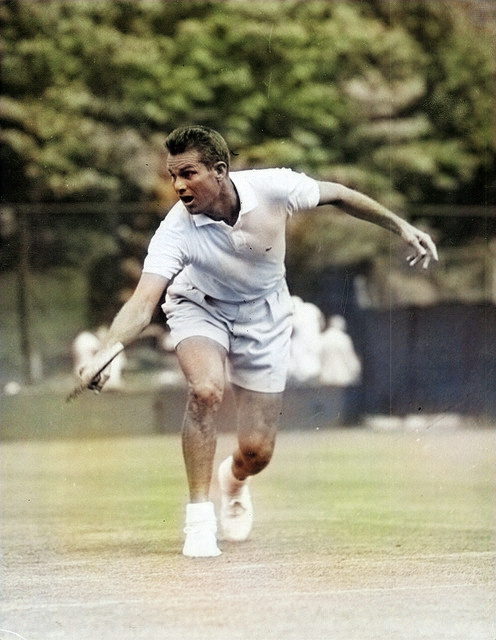}
& 
    \includegraphics[width=0.2\textwidth]{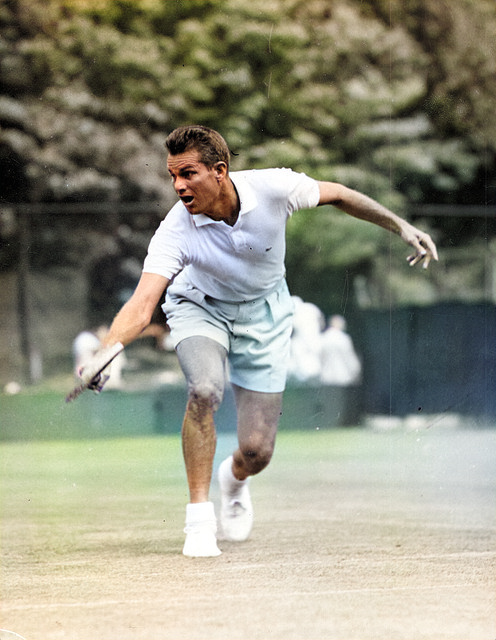}
\\
    \includegraphics[width=0.2\textwidth]{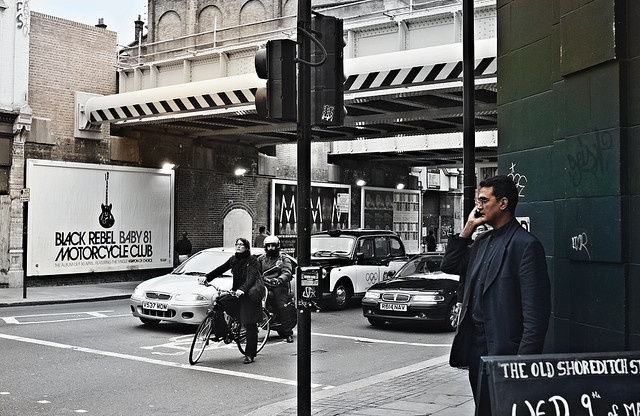}
& 
    \includegraphics[width=0.2\textwidth]{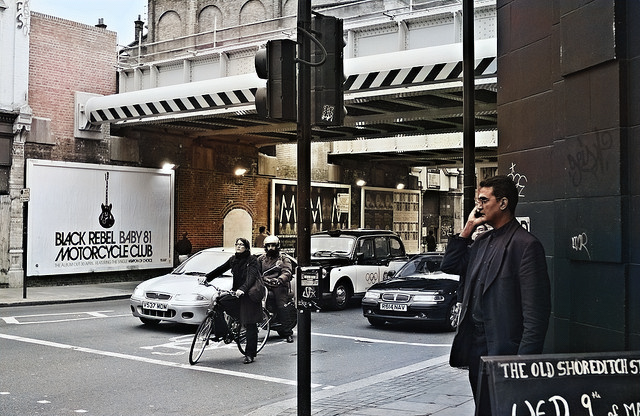}
& 
    \includegraphics[width=0.2\textwidth]{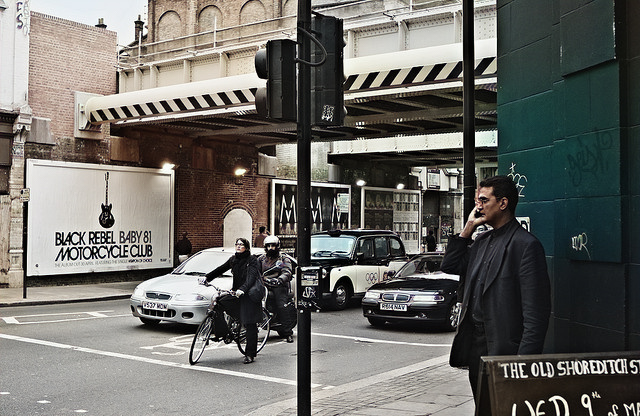}
& 
    \includegraphics[width=0.2\textwidth]{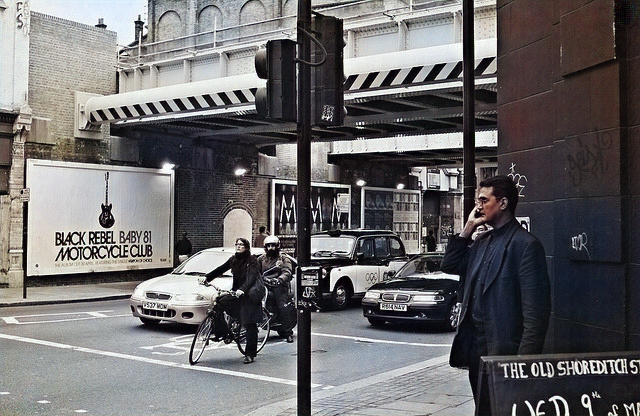}
& 
    \includegraphics[width=0.2\textwidth]{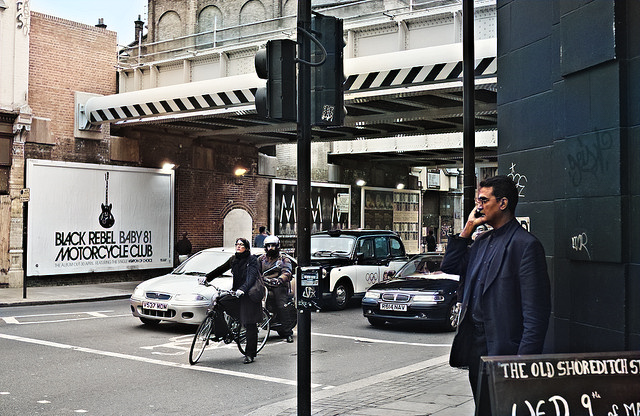}
\\
    \includegraphics[width=0.2\textwidth]{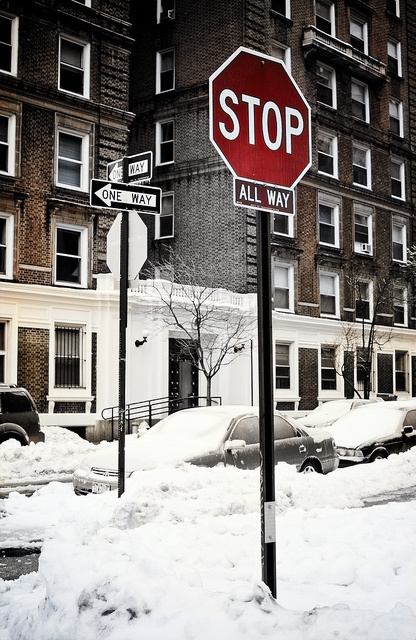}
& 
    \includegraphics[width=0.2\textwidth]{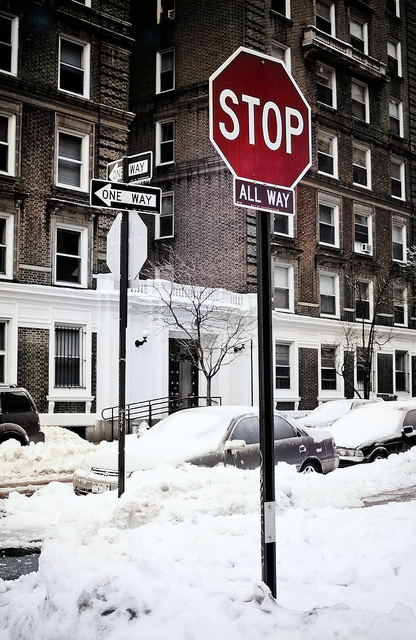}
& 
    \includegraphics[width=0.2\textwidth]{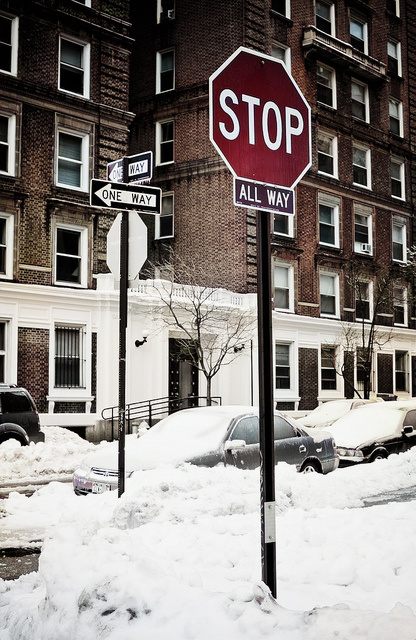}
& 
    \includegraphics[width=0.2\textwidth]{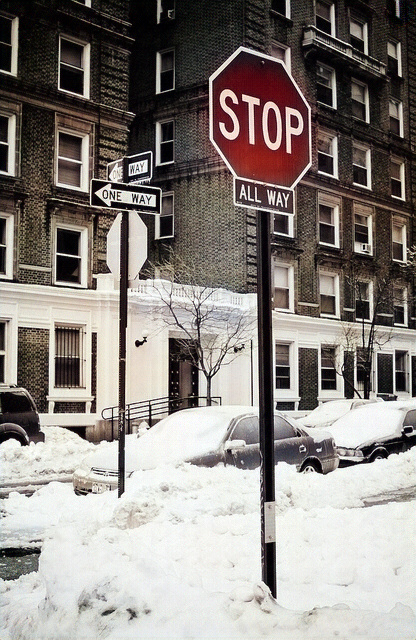}
& 
    \includegraphics[width=0.2\textwidth]{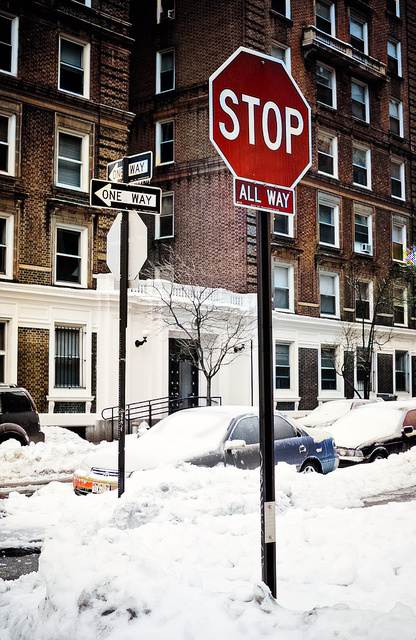}
\\
RGB-L1  & 
RGB-L2  & 
RGB-LPIPS &
RGB-WGAN+L2 &
RGB-WGAN+LPIPS
\\
\end{tabular}
\caption{Examples in Original Black and White Images. These colorization results have been obtained using the five networks trained, respectively, with  L1, L2, LPIPS, WGAN+L2 and WGAN+LPIPS losses, and learning the three RGB color channels.}
\label{oldRGB}
\end{center}
\end{figure}

Finally, in Figures~\ref{oldLab} and \ref{oldRGB}, we can see additional colorization results on real black and white images from the Pascal VOC dataset.
Those results have been obtained using the network trained with the five different losses, respectively, with L1, L2, VGG-based LPIPS, WGAN+L2, and WGAN+VGG-based LPIPS.
For Figure~\ref{oldLab}, only the two ab chrominance channels are learned while in Figure~\ref{oldRGB} the three RGB color channels are learned.
Again, none of the losses manage to consistently colorize the skin of all the people of the image at the first, second, and fourth rows of Figure~\ref{oldLab}, although possibly it is slightly better when using perceptual and GAN losses.
Notice that also in these cases, the colors are slightly more vivid, specially visible in the first two rows of Figure~\ref{oldLab}.
However, color inconsistency and failures in spatial localization appear, more visible in the first four rows.
As mentioned, this effect can be reduced by introducing semantic information (\emph{e.g.},~\cite{vitoria2020chromagan}) or spatial localization (\emph{e.g.},~\cite{su2020instance}).

\section{Conclusion}
\label{sec:conclu}

In this chapter, we have studied the role of loss functions on automatic colorization with deep learning methods.
Using a fixed standard network, we have shown that the choice of the right loss does not seem to play a crucial role in the colorization results.
We therefore argue that most efforts should be made on the influence of the architecture design, as it is related to the type of colorization operator one can expect to obtain.
Indeed, in our analysis, we used a U-Net based architecture which has shown to have a strong impact on the experimental results.
For the employed architecture, the models including the VGG-based LPIPS loss function provide overall slightly better results, especially for the perceptual metrics LPIPS and FID.
Likewise, the role of both architectures and losses for obtaining a real diversity of colorization results could be explored in future works.

\section*{Acknowledgements}

This study has been carried out with financial support from the French Research Agency through the PostProdLEAP project (ANR-19-CE23-0027-01) and from the EU Horizon 2020 research and innovation programme NoMADS (Marie Skłodowska-Curie grant agreement No 777826).
The first and fourth authors acknowledge partial support by MICINN/FEDER UE project, ref. PGC2018-098625-B-I00, and  RED2018-102511-T.
This chapter was written together with another chapter, called \textit{Influence of Color Spaces for Deep Learning Image Colorization} \cite{ballester2022influence}.
All authors have contributed to both chapters.

\bibliographystyle{apalike}
\bibliography{chapter_loss}

\end{document}